\documentclass[twocolumn,journal,vlined,ruled,linesnumbered]{IEEEtran}
\usepackage[T1]{fontenc}
\usepackage[latin9]{inputenc}
\usepackage{color}
\usepackage{array}
\usepackage{pifont}
\usepackage{textcomp}
\usepackage{multirow}
\usepackage{algorithm2e}
\usepackage{amsmath}
\usepackage{amssymb}
\usepackage{graphicx}
\usepackage{wasysym}
\usepackage[unicode=true,
 bookmarks=true,bookmarksnumbered=true,bookmarksopen=true,bookmarksopenlevel=1,
 breaklinks=false,pdfborder={0 0 0},pdfborderstyle={},backref=false,colorlinks=true]
 {hyperref}
\hypersetup{pdftitle={Your Title},
 pdfauthor={Your Name},
 pdfpagelayout=OneColumn, pdfnewwindow=true, pdfstartview=XYZ, plainpages=false}

\makeatletter

\providecommand{\tabularnewline}{\\}

\let\oldforeign@language\foreign@language
\DeclareRobustCommand{\foreign@language}[1]{%
  \lowercase{\oldforeign@language{#1}}}

\usepackage[caption=false,font=footnotesize]{subfig}
\usepackage{color}
\usepackage{calc}

\usepackage[switch]{lineno}

\@ifundefined{showcaptionsetup}{}{%
 \PassOptionsToPackage{caption=false}{subfig}}
\usepackage{subfig}
\makeatother

\begin{document}
\title{Global and Local Texture Randomization for Synthetic-to-Real Semantic
Segmentation}
\author{Duo~Peng,~Yinjie~Lei,~Lingqiao~Liu,~Pingping~Zhang~and~Jun~Liu
\thanks{This work was supported in part by Key Research and Development Program of Sichuan Province under Grant 2019YFG0409 and Fundamental Research Funds for the Central Universities under Grant DUT20RC(3)083. The associate editor coordinating the review of this manuscript and approving it for publication was Dr. Jiaying Liu. (\textit{Corresponding author: Yinjie Lei.})}
\thanks{Duo Peng and Yinjie Lei are with the College of Electronics and Information Engineering, Sichuan University, Chengdu 610064, China (e-mail: duo\_peng@stu.scu.edu.cn~and~yinjie@scu.edu.cn).}\thanks{Lingqiao Liu is with the School of Computer Science, University of Adelaide, Adelaide, SA 5005, Australia (e-mail: lingqiao.liu@adelaide.edu.au).}\thanks{Pingping Zhang is with the School of Artificial Intelligence, Dalian University of Technology, Dalian 116024, China (e-mail: zhpp@dlut.edu.cn).}
\thanks{Jun Liu is with the Information Systems Technology and Design Pillar, Singapore University of Technology and Design, Singapore (e-mail: jun\_liu@sutd.edu.sg).}}
\markboth{IEEE Transactions on Image Processing}{ Peng\MakeLowercase{\emph{et al.}}: Dr}
\maketitle
\begin{abstract}
Semantic segmentation is a crucial image understanding task, where each pixel of image is categorized into a corresponding label. Since the pixel-wise labeling for ground-truth is tedious and labor intensive,
in practical applications, many works exploit the synthetic images to train the model for real-word image semantic segmentation, i.e., Synthetic-to-Real Semantic Segmentation (SRSS). However, Deep Convolutional Neural Networks (CNNs) trained on the source synthetic data may not generalize well to the target real-world data. To address this problem,
there has been rapidly growing interest in Domain Adaption technique to mitigate the domain mismatch between the synthetic and real-world images. Besides, Domain Generalization technique is another solution to handle SRSS. In contrast to Domain Adaption, Domain Generalization seeks to address SRSS without accessing any data of the target domain during training. In this work, we propose two simple yet effective texture randomization mechanisms, Global Texture Randomization (GTR) and Local Texture Randomization (LTR), for Domain Generalization based SRSS. GTR is proposed to randomize the texture of source images into diverse unreal texture styles. It aims to alleviate the reliance of the network on texture while promoting the learning of the domain-invariant cues. In addition, we find the texture difference is not always occurred in entire image and may only appear in some
local areas. Therefore, we further propose a LTR mechanism to generate diverse local regions for partially stylizing the source images. Finally, we implement a regularization of Consistency between GTR and LTR (CGL) aiming to harmonize the two proposed mechanisms during training. Extensive experiments on five publicly available datasets (i.e., GTA5, SYNTHIA, Cityscapes, BDDS and Mapillary) with various SRSS settings (i.e., GTA5/SYNTHIA to Cityscapes/BDDS/Mapillary) demonstrate that the proposed method is superior to the state-of-the-art methods for domain generalization based SRSS.
\end{abstract}


\IEEEpeerreviewmaketitle{}

\section{Introduction\label{sec:Introduction}}

\begin{figure}[t]
\begin{centering}
\subfloat[Model training in Domain Adaptation.\label{fig:Domain-Adaption}]{\noindent \begin{centering}
\includegraphics[scale=0.56]{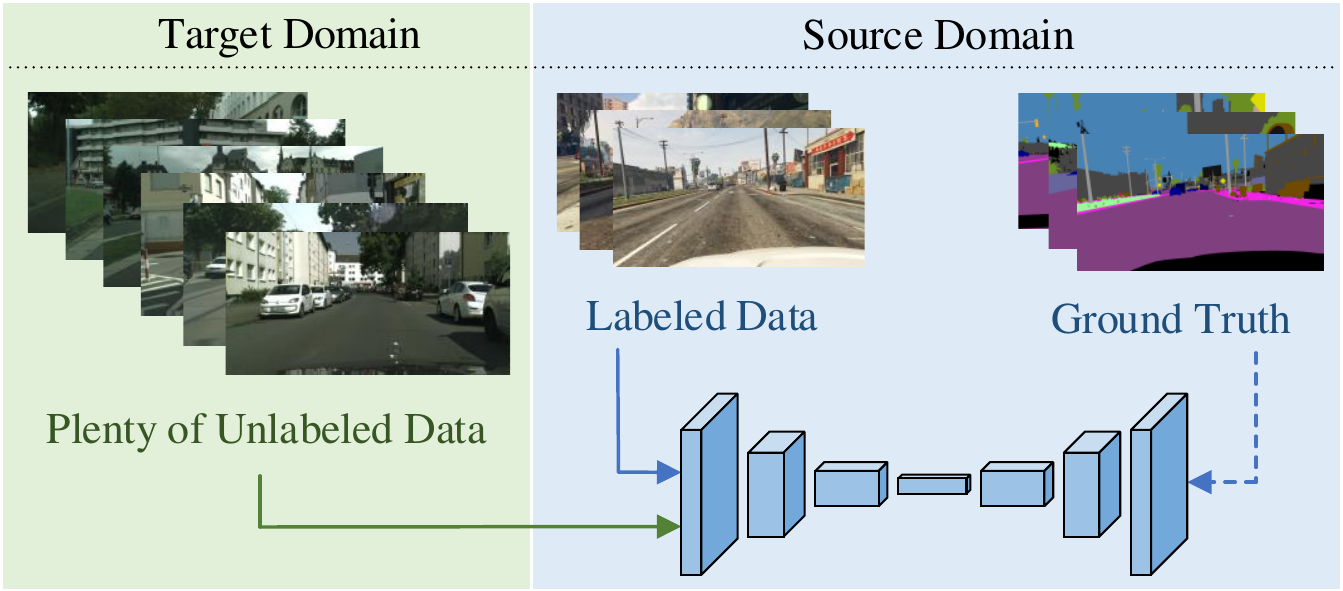}
\par\end{centering}
}
\par\end{centering}
\noindent \begin{centering}
\vspace{-2mm}
\par\end{centering}
\noindent \begin{centering}
\subfloat[Model training in Domain Generalization.\label{fig:Domain-Generalization}]{\noindent \begin{centering}
\includegraphics[scale=0.56]{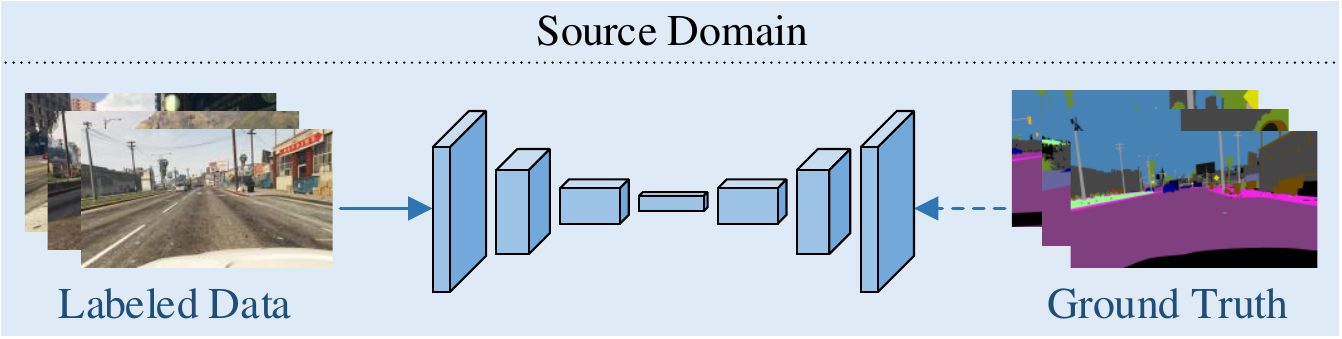}
\par\end{centering}
}
\par\end{centering}
\noindent \centering{}
\vspace{-2mm}
\caption{Comparison of the model training between Domain Adaptation and Domain
Generalization. Subfigure a: Domain Adaptation, which assumes that
the target domain is known beforehand and utilizes the unlabeled data
of target domain during training. Subfigure b: Domain Generalization
only uses the labeled images from source domain for training.}
\vspace{-5mm}
\end{figure}

\IEEEPARstart{S}{}emantic segmentation is a crucial computer vision
task for various practical applications, such as robotic navigation
and self-driving system \cite{zhang2019cascaded,long2015fully,chen2017deeplab,he2017mask}.
For semantic segmentation, pixel-wise labeling is
expensive and time consuming. Therefore, exploring economical methods
for automatically generating annotations is appealing, i.e., using
synthetic data whose ground-truth is straightforwardly provided \cite{handa2015scenenet,richter2016playing}.
To this end, extensive works have been conducted for \textbf{Synthetic-to-Real
Semantic Segmentation (SRSS)}. However, the learned model is often
challenged in practical scenarios where exists large difference between
source domain (synthetic images) and target domain (real-world images).
To solve this problem, \textbf{Domain Adaptation (DA)} \cite{hoffman2017cycada,hoffman2016fcns,zhang2017curriculum}\textbf{
}has recently been a research hot-spot. As shown in Fig. \ref{fig:Domain-Adaption},
it aims to narrow the domain gap by introducing the unlabeled data
from the target domain during training. However, data from the target
domain is not always accessible beforehand in practical scenarios.
Take autonomous driving for example, it is almost impossible to know
the vehicle will under which situation in advance (e.g., the unknown
weather, unknown city and unknown season). Based on the observations
above, another growing trend of SRSS is \textbf{Domain Generalization (DG)
}\cite{li2018learning,balaji2018metareg,ghifary2015domain,tobin2017domain}.
As shown in Fig. \ref{fig:Domain-Generalization}, DG is aimed at training without access to any target domain data.

In this work, we propose\textcolor{red}{{} }two texture randomization mechanisms, one consistency regularization and a painting selection strategy to handle the Domain Generalization based SRSS. More specifically, such two mechanisms are proposed to generate augmented images by merging various texture appearances from paintings, i.e., \textbf{Global Texture Randomization (GTR)} and \textbf{Local Texture Randomization (LTR)}. Next, we train a deep CNN model for semantic segmentation with \textbf{Consistency between GTR and LTR (CGL) }to harmonize the proposed two randomization mechanisms. Besides, a strategy namely\textbf{ Texture Complexity} \textbf{based Painting Selection (TCPS)} is proposed to ensure the selected paintings are reliable enough for the above texture randomization mechanisms. The next will be followed by a motivation on GTR, LTR, CGL and TCPS.

\begin{figure}[t]
\begin{centering}
\includegraphics[scale=0.17]{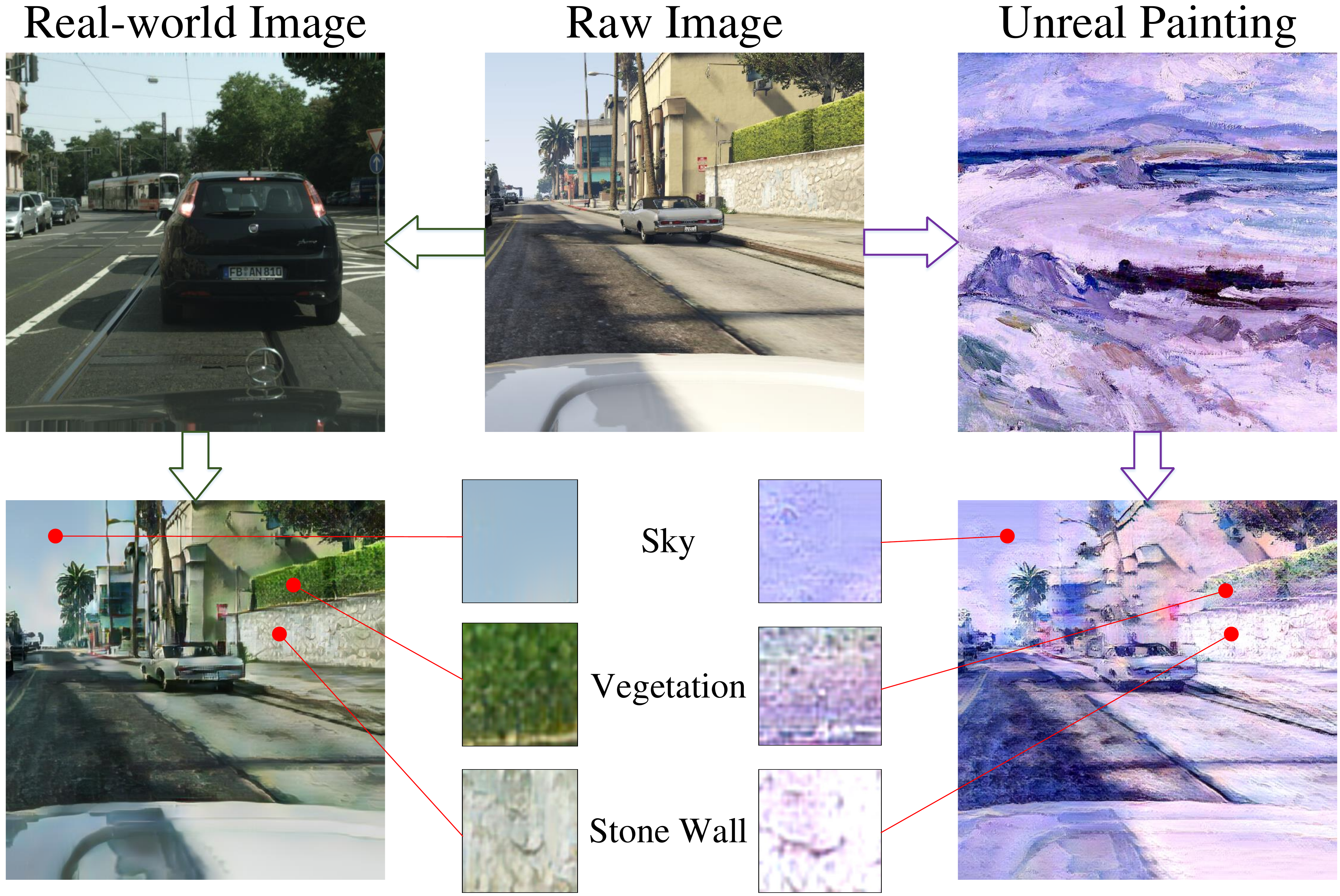}
\par\end{centering}
\vspace{-2mm}
\caption{Comparison between style transfer with real image and unreal painting.
\label{fig:real-unreal-GTR}}
\vspace{-6mm}
\end{figure}

\begin{figure}[t]
\begin{centering}
\includegraphics[scale=0.34]{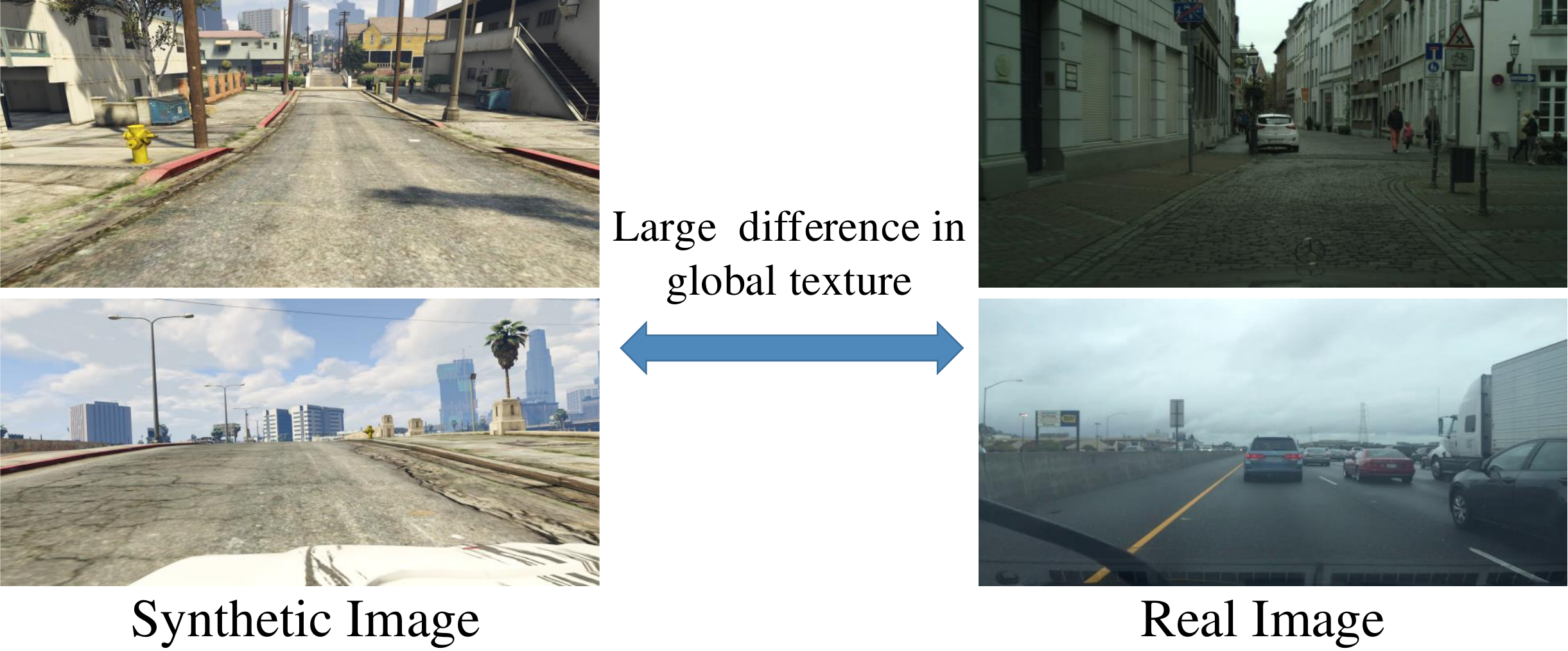}
\par\end{centering}
\vspace{-2mm}
\caption{Examples of images with large texture difference. It can be seen that the textures between the synthetic image and real image are globally different. \label{fig:Examples-of-images}}
\vspace{-6mm}
\end{figure}

\textbf{Global Texture Randomization (GTR).} Geirhos et al. \cite{GeirhosImageNet} has proved that CNNs are strongly biased towards the texture of training images. However, the texture appearances between the synthetic and real-world image are significantly different (see Fig. \ref{fig:Examples-of-images}), resulting in unsatisfactory SRSS performances. Existing methods \cite{bousmalis2017unsupervised,hoffman2017cycada} take advantages of such network texture bias by transferring the style of source images to that of \textbf{real images}. As shown in Fig. \ref{fig:real-unreal-GTR}, the real image can transfer the raw image into the domain of real-world scenario where the texture is regular and human-identifiable. However, when data from target domain is not accessible, these methods aimlessly seek various real images with diverse styles. This leads to inconvenience due to styles of real images should be manually selected in order to cover various possible scenarios that may occur in the target domain.

\begin{figure*}[tbh]
\begin{centering}
\includegraphics[scale=0.65]{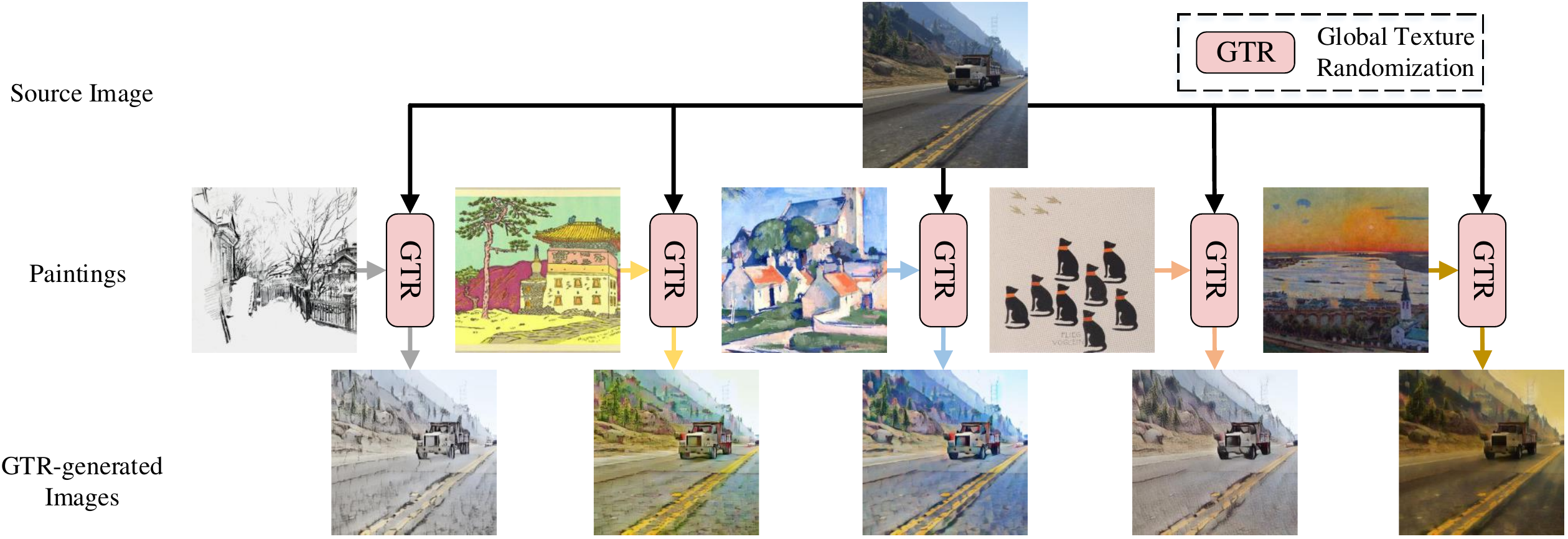}
\par\end{centering}
\vspace{-2mm}
\begin{centering}
\caption{Global Texture Randomization. Top: a cropped source image. Mid: texture images composed of paintings with various styles. Bottom: GTR-generated images with the same content information of source image and identical textures of paintings. \label{fig:Texture-Randomization}}
\par\end{centering}
\vspace{-2mm}
\end{figure*}

On the contrary, our approach leverages \textbf{unreal paintings} to handle style transfer. In the training phase, GTR is utilized to introduce more unreal textures which are random and irregular (shown in Fig. \ref{fig:Texture-Randomization}), making the difficulty of learning on image texture. Therefore, the network turns to learn more domain-invariant cues (e.g. shape and spatial layouts). As shown in Fig. \ref{fig:real-unreal-GTR}, compared to transfer with real image, using unreal paintings can make the transferred texture irregular and indistinct. Since the unreal texture is utilized to promote domain invariant learning instead of introducing new domains, the style of painting does not matter. The introduction of GTR-generated images essentially alleviate the network heavy reliance on image textures, making the model more robust to the global texture changes caused by the synthetic-to-real shift. To the best of our knowledge, this is one of the early works applying unreal textures to model training for SRSS.

\begin{figure*}[t]
\begin{centering}
\includegraphics[scale=0.37]{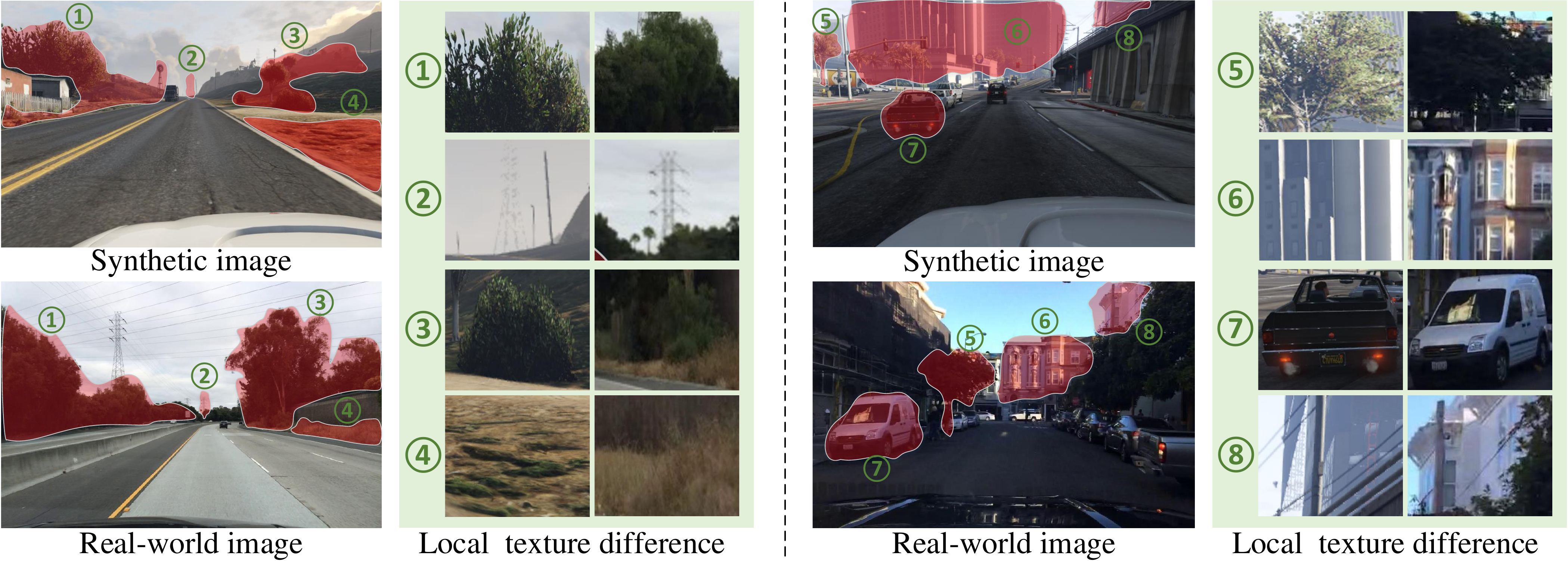}
\par\end{centering}
\vspace{-2mm}
\caption{Examples of images with local texture difference. It can be observed that only local areas between synthetic and real images are texture-different, especially for vegetation, terrain and building. \label{fig:3}}
\vspace{-6mm}
\end{figure*}

\begin{figure*}[tbh]
\begin{centering}
\includegraphics[scale=0.65]{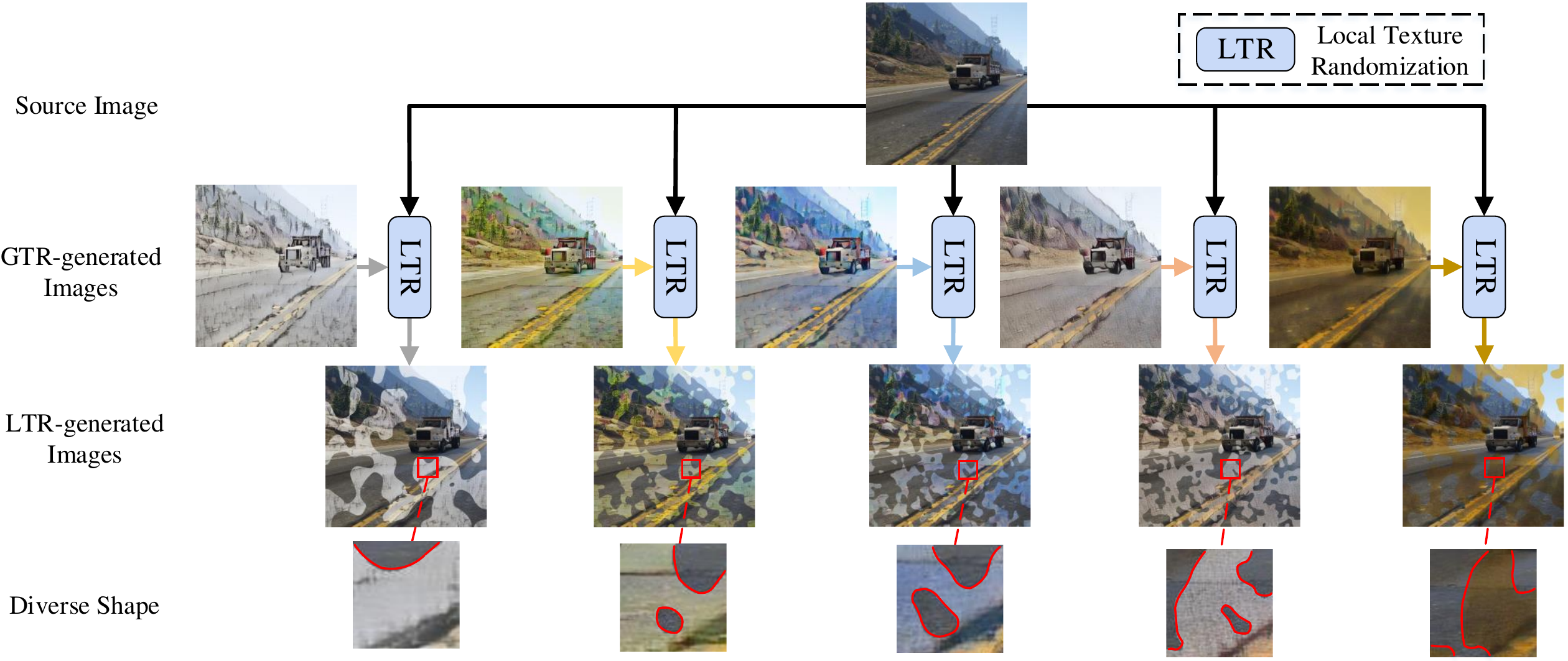}
\par\end{centering}
\caption{Local Texture Randomization. Aiming to obtain LTR-generated images (shown in the third row) which contains randomization regions with diverse shapes (shown in the last row), we mix the source image (shown in the first row) and each GTR-generated image (shown in the second row) respectively.\label{fig:Boundary-Randomization}}
\vspace{-5mm}
\end{figure*}

\textbf{Local Texture Randomization (LTR). }GTR tends to make the network completely ignore the texture information, which is mixed blessing: it indeed strengthens the model robustness to texture shift but also prevent the network from utilizing useful texture cues for segmentation. In fact, in most cases, only some local areas of the image are with texture differences. As shown in Fig. \ref{fig:3}, only the areas masked in red color contain large texture difference, while other areas are with small texture discrepancy. Thus, we propose a Local Texture Randomization (LTR) to address this situation. More specifically, LTR mechanism is proposed to generate images with local randomized texture for training. Since such local areas are usually with arbitrary shapes, LTR generates the local texture randomization areas with random boundaries (shown in Fig. \ref{fig:Boundary-Randomization}). Therefore, the usage of LTR enables CNNs to achieve a favorable performance under various local texture-different scenarios.

\textbf{Consistency between GTR and LTR (CGL).} In the proposed framework, both GTR and LTR are utilized for source image texture randomization. Nevertheless, source images simultaneously using both global and local texture randomization mechanisms may drive the network toward distinct learning directions and fail to convergence. To solve this problem, we propose a CGL mechanism to harmonize the proposed global and local texture randomization mechanisms during the training phase.

\textbf{Texture Complexity based Painting Selection (TCPS).} Since unreal textures are required to randomize source images, we utilize the paintings from dataset ``Painter by Numbers\textquotedblright \footnote{https://www.kaggle.com/c/painter-by-numbers/}. However, we observe that paintings with different texture complexities can significantly affect the model performance, while different style paintings with a same texture complexity lead to similar performances. In other words, the SRSS performance is sensitive to texture complexity instead of style of unreal paintings. Therefore it is necessary to give a measurement method for texture complexity so as to experimentally find what extent of texture complexity is suitable to SRSS. To this end, we propose a Texture Complexity based Painting Selection (TCPS) strategy to measure texture complexity and select reliable paintings.

The contributions of this paper are summarized as follows:
\begin{itemize}
\item We propose to utilize GTR mechanism to randomly replace the texture of source images with unreal textures, aiming to alleviate the network's reliance on texture.
\item We propose a LTR mechanism to further cover the cases in which only some local areas contain large texture difference between the synthetic and real images.
\item Moreover, we propose a consistency regularization between GTR and LTR to harmonize the proposed global and local texture randomization mechanisms during training.
\item Besides, we propose a painting selection strategy based on measurement of texture complexity to select reliable paintings for texture randomization.
\item The proposed method is comprehensively evaluated under various SRSS settings (i.e., GTA5/SYNTHIA to Cityscapes/BDDS/Mapillary), obtaining superior segmentation performance over other state-of-the-arts.
\end{itemize}

\section{Related Works}

\subsection{Semantic Segmentation\label{subsec:Semantic-Segmentation}}

Semantic segmentation is a crucial computer vision task and has attracted growing research interest as its various real-world applications. Before the advent of deep learning, traditional optimization techniques such as graph cuts \cite{kolmogorov2004energy}, belief propagation \cite{meltzer2005globally}, message passing \cite{kolmogorov2014new}, and random walks \cite{grady2004multi} have been widely used in semantic segmentation. For instance, \cite{sinop2007seeded} proposes a common framework which unifies graph cuts and random walks. \cite{dong2015sub} proposes a sub-Markov random walk algorithm to solve the problem of objects with thin and elongated parts. However, the energy formulations for describing natural images are complicated. Therefore higher-order energy functions \cite{blake2011markov} have been attracted more attention. \cite{shen2019submodular} proposes a framework of maximizing quadratic submodular energy with a more general knapsack constraint. Besides, an energy minimization method for general explicit higher-order energy functions is proposed \cite{shen2017higher}. It is competent in image segmentation with the appearance entropy.

As for deep learning based methods, the most frequently-used framework is Fully Convolutional Network (FCN) \cite{long2015fully}. It is capable of constructing an end-to-end mapping between the input RGB image and the semantic pixels. Afterwards, dilated convolution \cite{yu2015multi} has been proposed to modify the original FCN. Meanwhile, Markov and conditional random fields \cite{chen2014semantic,liu2015semantic,zheng2015conditional} are exploited to make up for the loss of object details. Afterwards, Global Convolutional Network \cite{peng2017large} is presented to use large kernels to handle semantic segmentation. \cite{pang2019towards} uses semantic enhancement module with feature attention for fusing of shallow features and deep features. Besides, Kernel-Sharing Atrous Convolution (KSAC) \cite{huang2019see} is proposed to share different parallel layers.

\subsection{Domain Adaptation\label{subsec:Domain-Adaption}}

It is known that, collecting a plenty of annotated training image data for each city of interest would be computationally expensive. To address this issue, an enormous amount of DA techniques are proposed. First of all, Richter \cite{richter2016playing} gave an exploration on extracting images from modern computer games with annotations generated by engine, inspiring a series of further studies. Most existing works fall into three categories: feature-level adaptation, pixel-level adaptation and both. (1) Feature-level adaptation aims to align features by either directly minimizing the feature distance \cite{chen2018road,long2016unsupervised,ganin2014unsupervised,cariucci2017autodial} or implicitly narrowing the gap between source and target distributions with adversarial learning \cite{hoffman2016fcns,chen2017no,tsai2018learning,tzeng2017adversarial,luo2019taking,vu2019advent,tsai2019domain}. (2) Pixel-level adaptation attempts to remove representation differences by stylizing source images to realistic target images \cite{zhu2017unpaired,murez2018image,bousmalis2017unsupervised,zou2018domain}. (3) The rest of methods take both levels described above into consideration \cite{hoffman2017cycada,wu2018dcan}. Even in a same adaptation level, methods varies at different alignment perspectives: class-wise alignment \cite{chen2017no,du2019ssf,luo2019taking}, patch-wise alignment \cite{tsai2019domain}, layer-wise alignment \cite{long2016unsupervised,cariucci2017autodial} and label-wise alignment \cite{tsai2018learning,zou2018domain}, etc.

\begin{figure*}[tbh]
\begin{centering}
\includegraphics[scale=0.55]{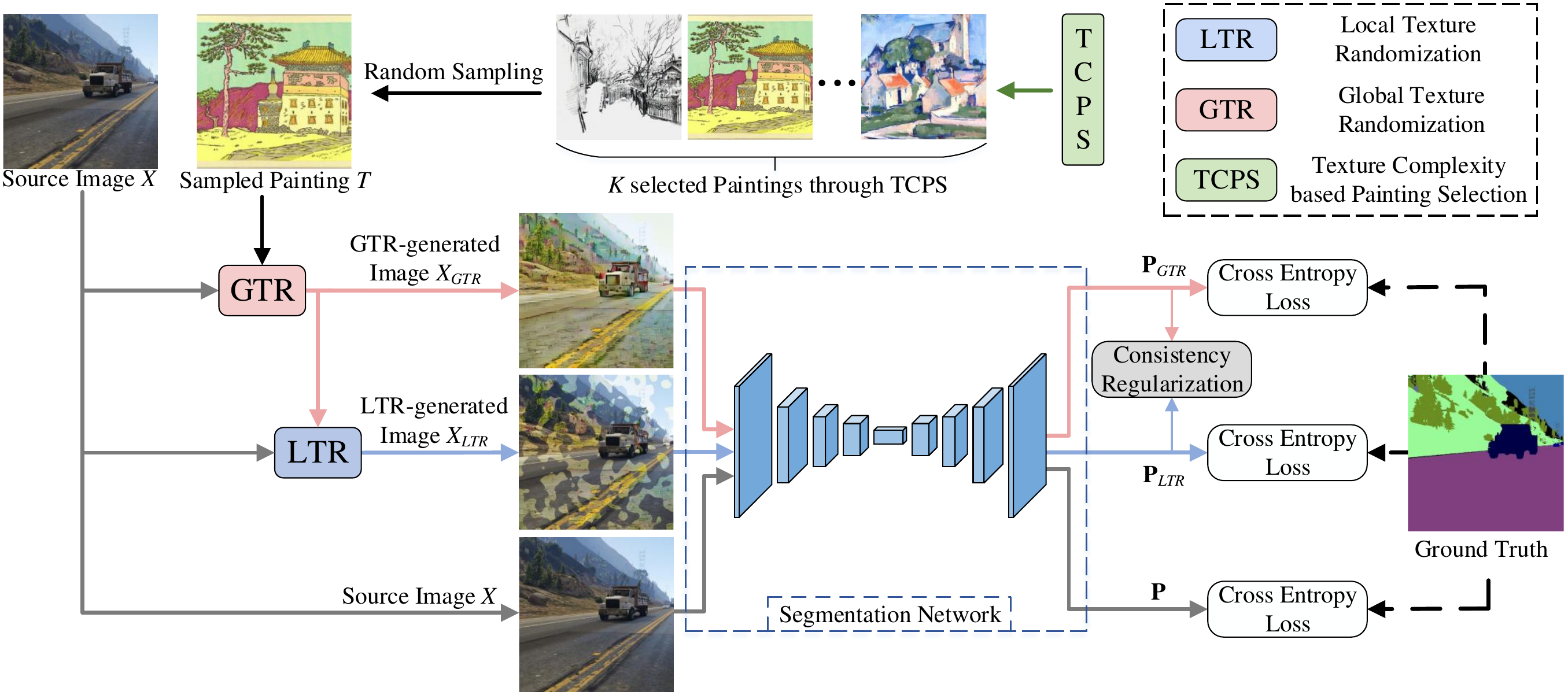}
\par\end{centering}
\vspace{-2mm}
\caption{The overall architecture of our framework. After TCPS, K paintings are obtained by random sampling. For each iteration, we randomly select one painting from $K$ paintings, namely $T$. With the help of the GTR/LTR mechanism, $X_{GTR}$/$X_{LTR}$ generated. There are 3 streams for network training. Besides the cross entropy loss on each stream, a consistency regularization is applied between GTR stream and LTR stream to harmonize the learning. \label{fig:Overview-of-the}}
\vspace{-5mm}
\end{figure*}

\subsection{Domain Generalization\label{subsec:Domain-Generalization}}

Unlike Domain Adaptation, Domain Generalization purely utilizes the labeled images in source domain \cite{muandet2013domain,gan2016learning}. \cite{gong2013reshaping} explores latent domains among training and testing data. \cite{li2018domain} presents a framework based on Maximum Mean Discrepancy (MMD) which measures the generalized latent feature representation across different domains. \cite{li2017deeper} develops a CNN model for an end-to-end Domain Generalization learning. \cite{li2018learning} utilizes the meta-learning strategy for the improvement of generalization performance. Although above methods are effective, the majority of them mainly focused on image classification. There are only a few DG studies addressing semantic segmentation \cite{yue2019domain,pan2018two,pan2019switchable}, which can be broadly classified as: layer-normalized generalization and data-enhanced generalization. (1) Layer-normalized generalization propose to search a reasonable combination of different whitening and standardization techniques in CNNs so as to improve generalization performance on unseen domains \cite{pan2018two,pan2019switchable}. (2) Data-enhanced generalization tends to randomize the synthetic images with various styles of real images, in order to effectively cover real-world target domain \cite{yue2019domain}. Although both \cite{pan2018two} and \cite{pan2019switchable} are layer normalized generalization, \cite{pan2019switchable} further takes whitening techniques (BW \cite{huang2018decorrelated} and IW \cite{li2017universal}) into consideration compared with \cite{pan2018two} that only utilizes standardization techniques (IN \cite{ulyanov2016instance} and BN \cite{ioffe2015batch}). However, training with various unreal features has been a largely under explored domain. In this paper, we propose to handle DG with randomized unreal images which implemented by Data Enhancement.

\subsection{Data Enhancement\label{subsec:Data-Enhancement}}

In this paper, we mainly provide a Data Enhancement method which resorts to Regularization Techniques and Data Augmentation.

\textbf{Data Augmentation.} Data Augmentation is the process of supplementing an extra dataset with similar data created from the raw information which is crucial in deep learning. When dealing with images, it often handles the image by rotating, cropping, blurring, adding noise and other operations to existing images, enabling the network to generalize better \cite{ciregan2012multi,sato2015apac,wan2013regularization,simard2003best}. \cite{lemley2017smart} automatically merges multiple images from the same class for data augmentation. A Bayesian based approach is proposed in \cite{tran2017bayesian} to augment the training set. \cite{devries2017dataset} applies simple transformations such as interpolating and extrapolating to perform the transformation in a learned feature space. In \cite{dreossi2018counterexample}, the misclassified examples are utilized for data augmentation.

\textbf{Regularization Techniques.} To generate randomized images, Regularization Techniques are also utilized. Hence, some common Regularization Techniques are introduced in this part. First, the MixUp regularization \cite{zhang2017mixup} improves the performance of network training by randomly interpolating samples from training set. Then, the Cutout regularization \cite{devries2017cutout} augments the training data by shielding a rectangular region in image. Recently, \cite{yun2019cutmix} presents CutMix regularization which combines both MixUp and Cutout. Due to the addition of a stronger perturbation, CutMix improves model performance and outperforms the other two methods (i.e., MixUp and Cutout).

\section{Methodology\label{sec:Methodology}}

\subsection{Overview of the Proposed Framework}

The overall architecture of our framework is shown in Fig. \ref{fig:Overview-of-the}. First, we utilize TCPS to execute pre-selection towards dataset ``Painter by Numbers\textquotedblright{} and obtain a subset of dataset. Then $K$ paintings are given by random sampling from subset. In this paper, the hyper-parameter $K$ is set to 15. Since our goal is to randomize the texture of source image during training, one painting $T$ is randomly selected from $15$ paintings in each iteration. Formally, we consider a source image $\{X\}\in\mathbb{R}^{H\times W\times3}$ along with a selected painting $T$. The image $X$ and painting $T$ are fed forward into GTR module to generate a stylized image $X_{GTR}$. Next, we simultaneously send $X$ and $X_{GTR}$ into LTR to generate a local stylized image $X_{LTR}$. Afterwards, $X$, $X_{GTR}$ and $X_{LTR}$ are fed into the segmentation network to obtain segmentation outputs\textbf{ $\mathbf{P}$}, $\mathbf{P}_{GTR}$ and $\mathbf{P}_{LTR}$ respectively. For a better convergence, consistency is applied between GTR and LTR to make segmentation predictions (i.e., $\mathbf{P}_{GTR}$ and $\mathbf{P}_{LTR}$) close to each other. Finally, each stream of segmentation losses is computed to update the network's weights. During testing, we remove these modules and only retain the segmentation network. 

\begin{figure}[t]
\begin{centering}
\centering{}\centering \resizebox{0.45\textwidth}{!}{\includegraphics[scale=0.25]{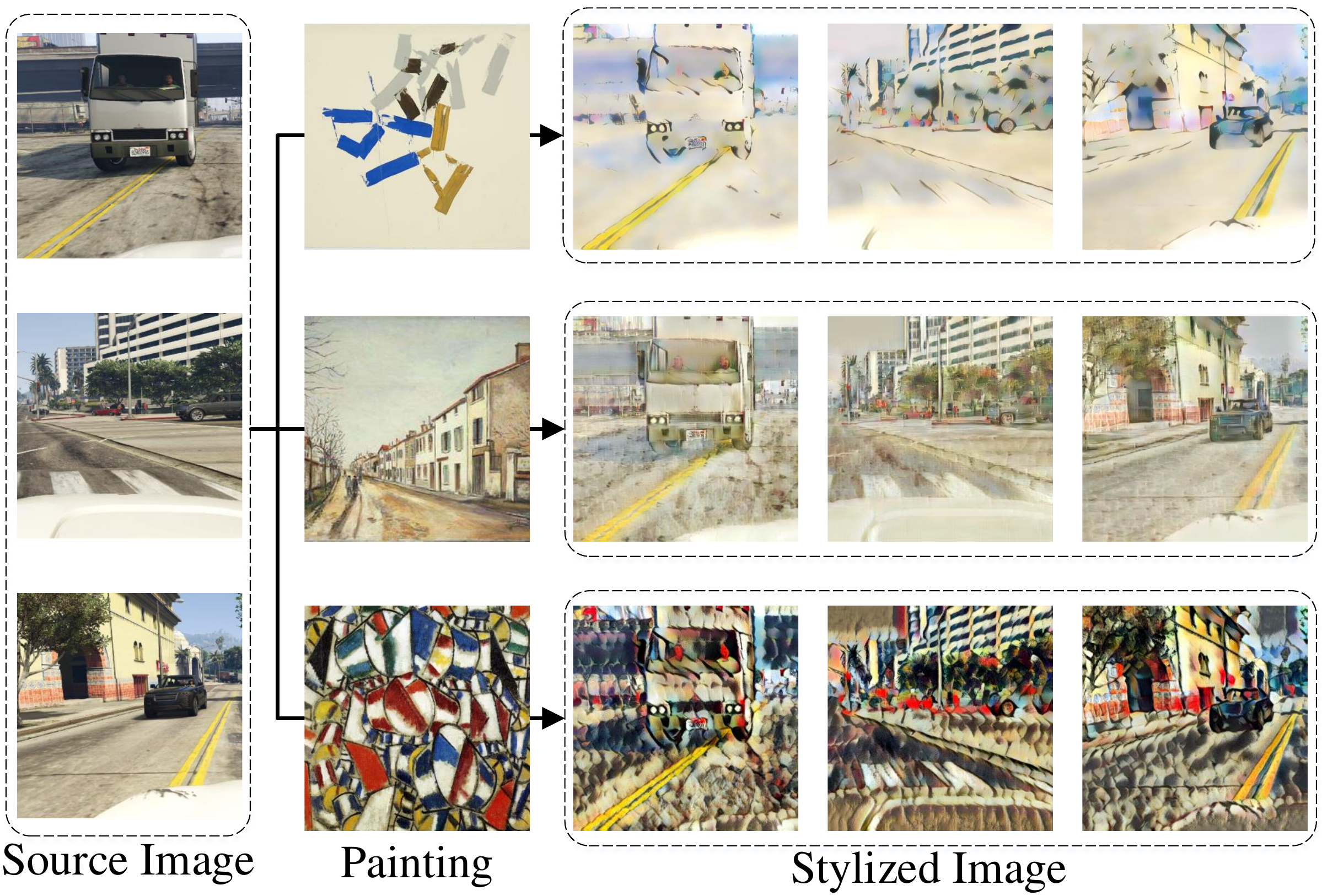}}
\par\end{centering}
\vspace{-4mm}
\caption{Texture randomization results using paintings with different texture complexities. The texture complexity of painting increases from the top to bottom. \label{fig:The-motivation-for-TAM}}
\vspace{-4mm}
\end{figure}

\subsection{Texture Complexity based Painting Selection (TCPS)}

From Fig. \ref{fig:The-motivation-for-TAM}, we can see that the stylized image performs unfavorably when texture complexity of the painting becomes quite high (i.e., the third row) or low (i.e., the first row). The key to ensure the generalization performance is to select paintings which can effectively replaces image texture while preserving the shape of objects. In light of this, we propose a Texture Complexity measurement to select paintings from dataset ``Painter by Numbers\textquotedblright .

According to the gradient representation of each pixel, we divide an image into two parts: (1) smooth area, where pixel values change slightly or have no change; (2) unsmooth area, where pixel values change significantly. Inspired by Structure Tensor \cite{kothe2003edge}, we separate the two areas by calculation of gradient representation for each pixel. Assume $Grad(i,j)$ denotes the gradient representation of pixel at row $i$ and column $j$.

\begin{figure}[t]
\begin{centering}
\includegraphics[scale=0.30]{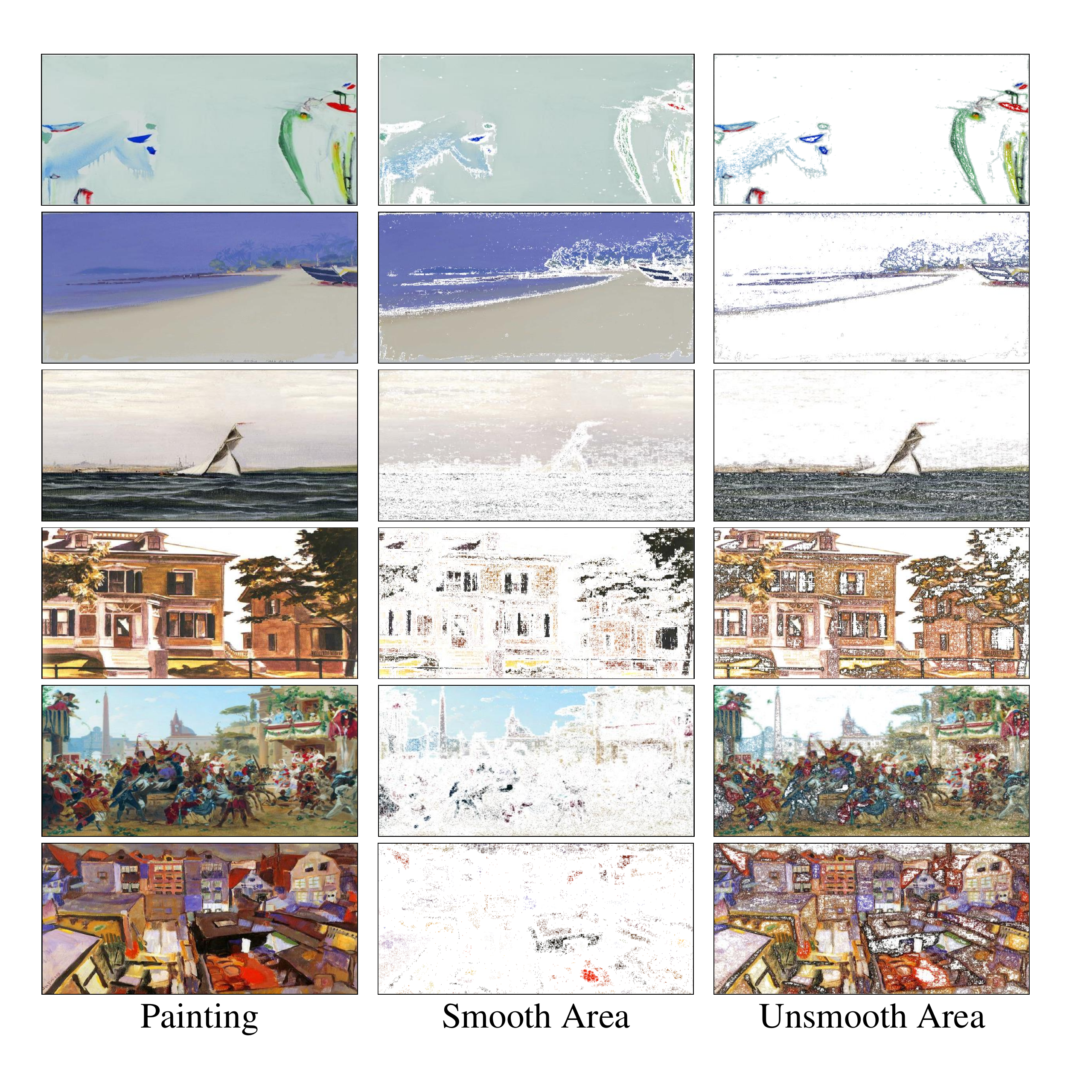}
\par\end{centering}
\vspace{-4mm}
\caption{Examples of smooth-unsmooth area separation from dataset ``Painter by Numbers\textquotedblright . The first column shows the raw paintings. The last two columns are the separated smooth area and the corresponding unsmooth area.\label{fig:The-results-of-TCAM}}
\vspace{-5mm}
\end{figure}

The gradient representation $Grad(i,j)$ can reflect the extent of change in pixel value and hence indicates whether the area of pixel is smooth or not. Pixels with little gradient representation are formed into the smooth area, such as sky, wall and background. On the contrary, pixels with large gradient representation make up the unsmooth area which has a distinct texture expression. To quantitatively determine which area the pixel belongs to, we set a threshold $\varepsilon$ manually. When meeting the condition: $Grad(i,j)<\varepsilon$, the pixel is allocated to the smooth area, or else it belongs to the unsmooth area. In this paper, we set $\varepsilon$ to 20. Fig. \ref{fig:The-results-of-TCAM} shows some examples of smooth-unsmooth area separation by using our method. Qualitative results demonstrate the effectiveness of our method.

As textures of an image almost gather in unsmooth area, the texture complexity can be represented by the proportion of unsmooth area in the image. It can be formulated as following:

\vspace{-2mm}

\begin{equation}
\mathcal{T_{\mathit{exc}}}=\frac{\underset{i}{\sum}\underset{j}{\sum}Bool(I(i.j)\in A_{usm})}{N}
\end{equation}
where $Bool(\cdot)$ is the judgment function, if the condition is true it returns 1, or else it returns 0; $A_{usm}$ denotes the unsmooth area; $N$ is the total number of pixels. In essence, the texture complexity is the proportion of unsmooth area in the image. By calculating the texture complexity, we can determine if this painting is competent to carry out texture randomization. Based on the ablation study in Section \ref{sec:Ablation-Study}, we find the painting with $\mathcal{T}_{exc}\in[0.55,0.65]$ can achieve an pleasurable and stable texture randomization performance.

\subsection{Global Texture Randomization (GTR)}

\begin{figure}[t]
\begin{centering}
\includegraphics[scale=0.30]{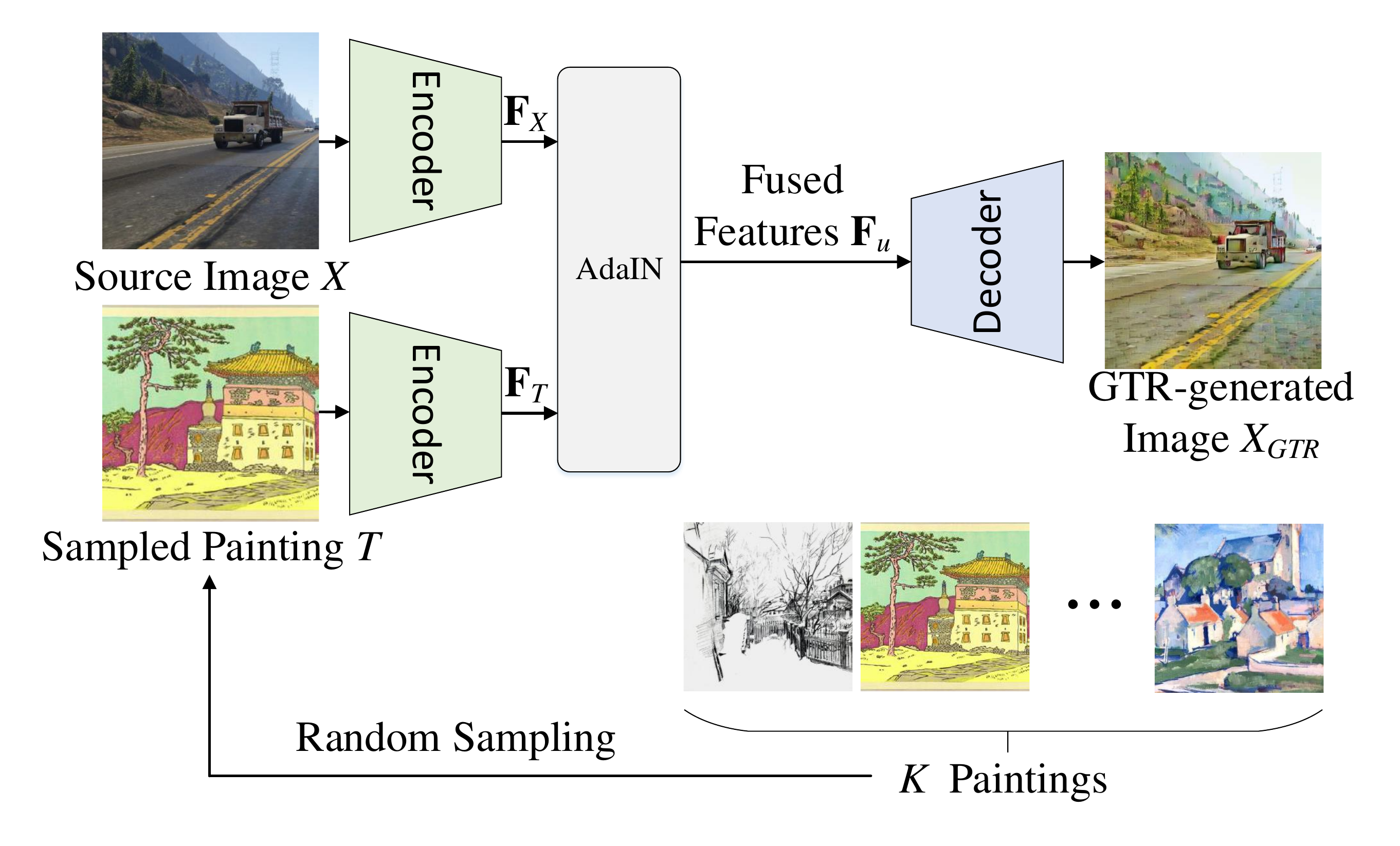}
\par\end{centering}
\vspace{-2mm}
\caption{The detailed architecture of the GTR module. The major structure of GTR module is a pre-trained style translation network \cite{huang2017arbitrary}. Aiming to replace the texture of the raw image $X$ to that of the painting $T$, an AdaIN layer \cite{huang2017arbitrary} is utilized to fuse the features extracted by the encoder. Then, the fused feature map $\mathbf{F}_{u}$ is fed forward to the decoder to obtain the GTR image $X_{GTR}$.\label{fig:The-detailed-architecture}}
\vspace{-4mm}
\end{figure}

GTR aims to strip the texture of the source image and replace with the style of the painting. Note that in each iteration, we implement random sampling on $K$ paintings to acquire the painting $T$. We then use the pre-trained style translation network \cite{huang2017arbitrary} to transfer the texture of the selected painting to the source image $X$. More specifically, As shown in Fig. \ref{fig:The-detailed-architecture}, the raw image $X$ and the painting $T$ are first fed into a pre-trained Siamese encoder for feature extraction. The feature extraction can be formulated as following:\vspace{-2mm}

\begin{equation}
\mathbf{F_{\mathrm{\mathit{X}}}}=f(X,\boldsymbol{\theta}^{en})
\end{equation}

\vspace{-2mm}

\begin{equation}
\mathbf{F_{\mathrm{\mathit{T}}}}=f(T,\boldsymbol{\theta}^{en}),
\end{equation}
where $\mathbf{F}_{X}$ and $\mathbf{F}_{T}$ denote the extracted feature maps from source image $X$ and painting $T$, $\boldsymbol{\theta}^{en}$ denotes the weights of the encoder network $f(\cdot,\cdot)$ .

Then an Adaptive Instance Normalization (AdaIN) \cite{huang2017arbitrary} is utilized to perform feature fusion. Particularly, AdaIN treats the mean and variance of the extracted feature maps (i.e., $\mathbf{F}_{X}$ and $\mathbf{F}_{T}$) as affine parameters. The feature fusion implemented by AdaIN can be written as following:

\begin{equation}
\mathbf{F_{\mathit{u}}}=\sigma(\mathbf{F_{\mathrm{\mathit{T}}}})\cdot(\frac{\mathbf{F_{\mathrm{\mathit{X}}}}-\mu(\mathbf{F_{\mathrm{\mathit{X}}}})}{\sigma(\mathbf{F_{\mathrm{\mathit{X}}}})})+\mu(\mathbf{F_{\mathrm{\mathit{T}}}}),
\end{equation}
where $\mathbf{F_{\mathit{u}}}$ is the fused feature map, $\mu(\cdot)$ and $\sigma(\cdot)$ are the mean and variance values which are spatially computed across the entire feature maps.

Finally, a pre-trained decoder inverts the transferred feature map $\mathbf{F_{\mathit{u}}}$ to the GTR image $X_{GTR}$. Formally, the GTR-generated image $X_{GTR}$ can be calculated by:

\vspace{-2mm}

\begin{equation}
X_{GTR}=f(\mathbf{F_{\mathit{u}}},\boldsymbol{\theta}^{de}),
\end{equation}
where $\boldsymbol{\theta}^{de}$ denotes the weights of the decoder network.

After the process of GTR module, the texture variety of training set is augmented, leading to robustness of global texture change. Despite such effectiveness, the model with GTR ignores the cases where only local texture difference exists between synthetic and real images. We improve it by the following Local Texture Randomization.

\subsection{Local Texture Randomization (LTR)}

\begin{figure}[t]
\begin{centering}
\includegraphics[scale=0.40]{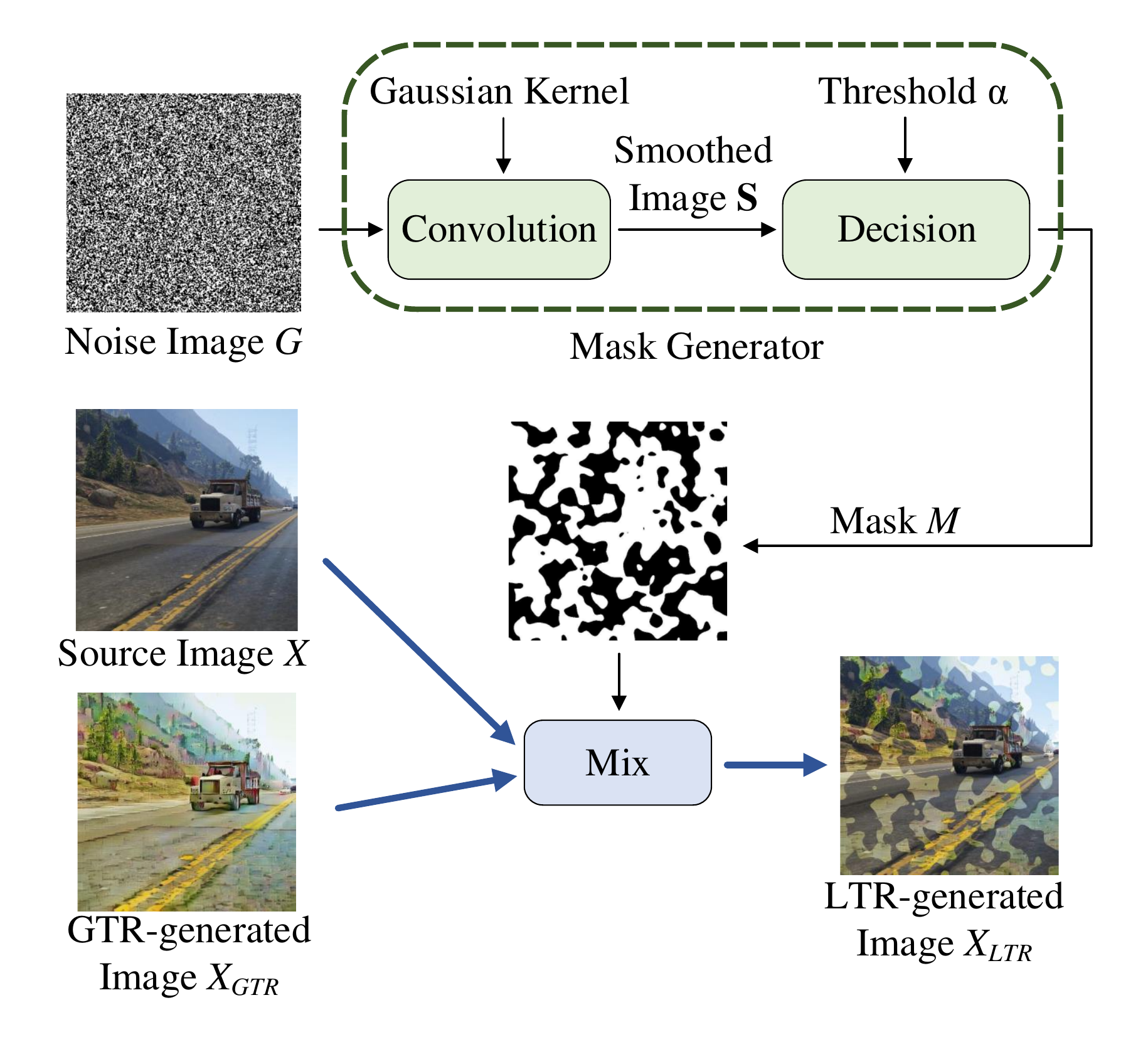}
\par\end{centering}
\vspace{-2mm}
\caption{Illustration of the proposed LTR module. To generate the mask, a well-designed mask generator \cite{french2019semi} is applied to a noise image.\label{fig:An-illustration-of}}
\vspace{-2mm}
\end{figure}

LTR is proposed to randomly generate images with local randomized textures. For LTR module, the raw image $X$ and GTR-generated image $X_{GTR}$ are fed into LTR module to generate the output image $X_{LTR}$. In each iteration, LTR generates a mask $M$ with a random shape to mix the source image and GTR-generated image so that the masks among different iterations are diverse. As shown in Fig. \ref{fig:An-illustration-of}, we utilize the mask generator \cite{french2019semi} to create mask $M$. For more details, we firstly generate an image with Gaussian noise which follows a normal distribution $\mathcal{N}(0,1)$, named as $G$. Then, the noise image $G$ is smoothed with a Gaussian kernel:

\begin{equation}
\mathbf{S}=f^{3\times3}(G,\gamma),
\end{equation}
where $\mathbf{S}$ denotes the smoothed image, $f^{3\times3}(\cdot)$ denotes the convolution with $3\times3$ Gaussian kernel and $\gamma$ is the standard deviation of Gaussian kernel. As for $\gamma$, we use a nonlinear calculation on variable parameter $\lambda$ to obtain it:

\vspace{-2mm}

\begin{equation}
\gamma=\exp(\lg(\lambda)).\label{eq:}
\end{equation}

\begin{figure}[t]
\begin{centering}
\includegraphics[scale=0.28]{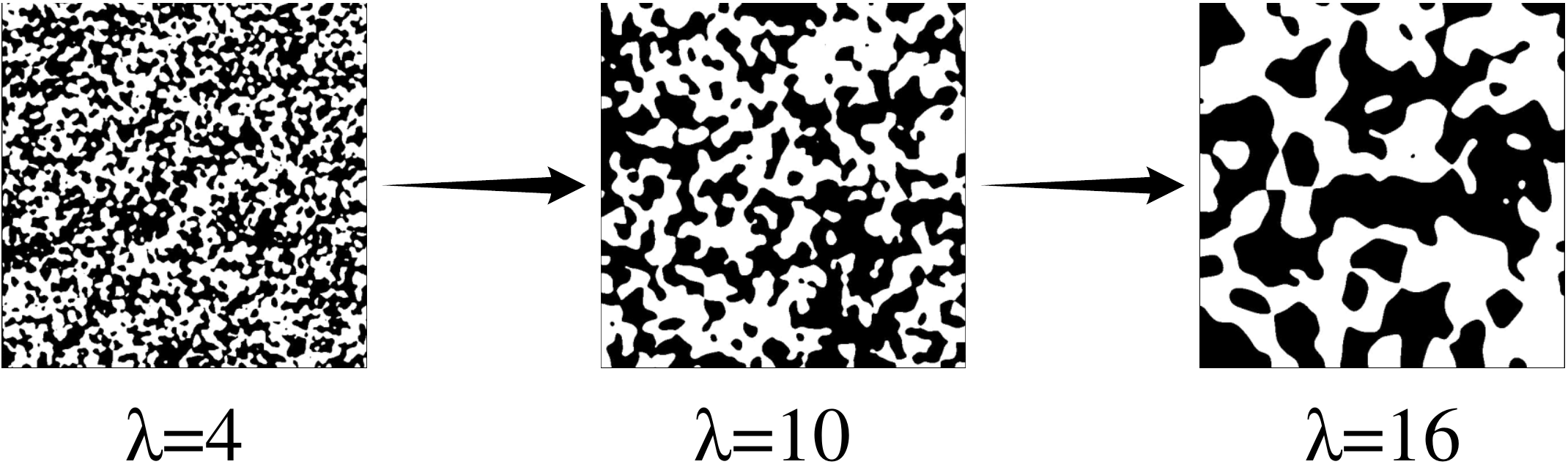}
\par\end{centering}
\vspace{-2mm}
\caption{Example of masks varying $\lambda$ from 4 to 16 with resolution 640\texttimes 640.\label{fig:Example-masks-varying}}
\vspace{-4mm}
\end{figure}

In each training iteration, we obtain $\lambda$ from the uniform distribution $[\lambda_{min},\lambda_{max}]$ by random selection. After ablation study in Section V-B, we respectively set $\lambda_{min}$ and $\lambda_{max}$ to 4 and 16 in this paper. In particular, the random determination of $\lambda$ leads to the variety of mask. Fig. \ref{fig:Example-masks-varying} shows some generated masks with varying values of $\lambda$.

At the stage of generating mask, a threshold $\alpha$ is used to decide whether each pixel is set to 0 (black) or 1 (white). For a mask which contains white region with a proportion of $p$, we can obtain the threshold $\alpha$ through ranking all pixel values and choosing the last value of former $p$ elements. However, ranking algorithm is computationally expensive, leading to slow training speed. To alleviate this problem, previous study \cite{french2019semi} has shown that this goal can be achieved through the error function \cite{andrews1998special} without ranking pixel values. Therefore, given a proportion $p$, we can directly compute the threshold $\alpha$ as following:

\vspace{-2mm}

\begin{equation}
\alpha=erf^{-1}(2p-1)\cdot\sqrt{2}\sigma_{S}+\mu_{S},
\end{equation}
where $erf^{-1}(\cdot)$ is the inverse function of error function $erf(\cdot)$ \cite{andrews1998special}, $\mu_{S}$ and $\sigma_{S}$ are the mean value and standard deviation of the smoothed image $\mathbf{S}$ respectively. Particularly, $p$ is the hyper parameter to decide the proportion of white region. After obtaining the threshold $\alpha$, we turn the smoothed image $\mathbf{S}$ into mask $M$. The $i$-th element $M_{i}$ of the mask $M$ can be calculated by:

\vspace{-2mm}

\begin{equation}
M_{i}=\begin{cases}
0 & E_{i}<\alpha\\
1 & E_{i}\geq\alpha
\end{cases},
\end{equation}
where $E_{i}$ denotes the $i$-th element in $\mathbf{S}$.

Finally, we mix the raw image $X$ and GTR-generated image $X_{GTR}$ with mask $M$. We define the mix function as $Mix(X,X_{GTR},M)$. Specifically, the LTR-generated image $X_{LTR}$ can be computed as:

\vspace{-2mm}

\begin{equation}
X_{LTR}=Mix(X,X_{GTR},M)=(1\text{\textminus}M)\odot X+M\odot X_{GTR},
\end{equation}
where $\odot$ denotes pixel-wise multiplication. With this step, $X_{LTR}$ is generated to further improve the generalization ability.

\subsection{Consistency between GTR and LTR (CGL)\label{subsec:Consistency-between-TR}}

Simply training the network with GTR images and LTR images has two weaknesses: (a) different randomization methods will fail to converge for network training; (b) even if the network converge well, the network may be trained towards completely different directions resulting in unstable performance. To this end, we impose the consistency loss to harmonizes the two randomization learning methods. $L1$ distance is utilized for the calculation of consistency loss. The formulation of proposed CGL can be written as:

\vspace{-2mm}

\begin{equation}
\mathcal{L}_{con}(X_{GTR},X_{LTR})=\mathcal{L}_{1}(f(X_{GTR},\boldsymbol{\theta}),f(X_{LTR},\boldsymbol{\theta})),
\end{equation}
where $\mathcal{L}_{1}(\cdot,\cdot)$ denotes the $L1$ distance, $\boldsymbol{\theta}$ is the weights of the whole segmentation network $f(\cdot,\cdot)$ . Note that we extract the feature maps from the last convolutional layer before softmax to calculate the $L1$ distance.

\subsection{Network Training}

Given a set of training samples $\mathcal{D}=\{X,X_{GTR},X_{LTR},Y\}$, where $X$, $X_{GTR}$ and $X_{LTR}$ are the input raw image, GTR-generated image and LTR-generated images, $Y$ is the ground-truth. Considering both GTR and LTR are utilized in training, we define the following segmentation loss:

\vspace{-2mm}

\begin{equation}
\mathcal{L}_{seg}=\mathcal{L}_{w}(X,Y)+\mathcal{L}_{w}(X_{GTR},Y)+\mathcal{L}_{w}(X_{LTR},Y),\label{eq:lossseg}
\end{equation}
where $\mathcal{L}_{w}(X,Y)$, $\mathcal{L}_{w}(X_{GTR},Y)$ and $\mathcal{L}_{w}(X_{LTR},Y)$ is the weighted cross entropy loss of raw image, GTR-generated image and LTR-generated image, respectively.

Considering the consistency regularization described in Section \ref{subsec:Consistency-between-TR}, our final training loss can be rewritten as:

\begin{equation}
\mathcal{L}=\mathbf{arg~min}~(1-\beta)\mathcal{L}_{seg}+\mathrm{\beta}\cdot\mathcal{L}_{con}(X_{GTR},X_{LTR}),\label{eq:beta}
\end{equation}
where $\beta$ is the hyper-parameter which trades off the quality of segmentation loss ($\mathcal{L}_{seg}$) with consistency loss ($\mathcal{L}_{con}$). In this paper, $\beta$ is set to 1e-5. As for testing, the image only pass through the segmentation network without the process of GTR, LTR and CGL.

\section{Experimental Setups\label{sec:Experimental-Setups}}

\subsection{Datasets Description}

\noindent \textbf{Source domain datasets:}

\textit{GTA5} is a synthetic image dataset collected in a computer game with pixel-wise semantic labels. It contains 24966 images with resolution of 1914 $\times$ 1052. It includes 19 classes which are compatible with most semantic segmentation datasets of outdoor scenes.

\textit{SYNTHIA} is a large synthetic dataset with pixel-level semantic annotations. The subset SYNTHIA-RANDCITYSCAPES is used in our experiments which contains 9400 images. It consists of 16 categories with high resolution of 1280 $\times$ 760.

\noindent \textbf{Target domain datasets:}

\textit{Cityscapes} contains vehicle-centric urban street images taken from some European cities. There are 5000 images with pixel-wise annotations. The images with size of 2048 $\times$ 1024 are labeled into 19 classes.

\textit{BDDS} contains thousands of real-world dashcam video frames with accurate pixel-wise annotations. It contains 34 labeled categories with resolution of 1280 $\times$ 720. The training, validation, and test sets contain 7000, 1000 and 2000 images, respectively.

\textit{Mapillary} contains street-level images with diverse resolutions. The annotations contain 66 object classes, but only 19 and 16 classes that overlap with GTA5 and SYNTHIA respectively, are used in our experiments. It has a training set with 18000 images and a validation set with 2000 images.

\noindent \textbf{SRSS experimental settings:}

\ding{172} GTA5 to Cityscapes, \ding{173} GTA5 to BDDS, \ding{174} GTA5 to Mapillary, \ding{175} SYNTHIA to Cityscapes, \ding{176} SYNTHIA to BDDS and \ding{177} SYNTHIA to Mapillary.

\subsection{Implementation Details\label{subsec:Implementation-Details}}

For fair comparison with other Domain Generalization methods, we use VGG-16 \cite{simonyan2014very}, ResNet-50 and ResNet-101 \cite{he2016deep} as the segmentation network respectively. The source codes and models are trained and evaluated on PyTorch toolbox \cite{paszke2019pytorch} based on Python 3.6 platform\footnote{The source codes will be released upon the acceptance of this paper.}. For more details, all proposed models are implemented on 8 NVIDIA RTX 2080Ti GPUs and two E5-2620 CPUs.

\textbf{Evaluation Metric.} Following previous works, we utilize PASCAL VOC Intersection over Union (IoU) \cite{everingham2015pascal} as the evaluation metric for testing. And mIoU is the mean value of IoUs across all categories. There is a positive correlation between mIoU value and segmentation performance.

\textbf{Data Preprocessing}. For the training set in GTA5, we crop each panoramic image into two patches with a resolution of 1052 $\times$ 1052. Then, we resize each cropped image into a resolution of 640 $\times$ 640. For the training data in SYNTHIA dataset, we also crop each image into two patches with a resolution of 640 $\times$ 640. Since downsampling of CNN, we restore the resolution of the output prediction by upsampling with Bi-linear Interpolation. Finally, as \cite{zhao2017pyramid,chen2019domain} do, we further adopt random mirror and Gaussian blur to images before network training.

\textbf{Parameter Settings}. The model is initialized with the network parameters pre-trained on ImageNet \cite{deng2009imagenet:} except those of the final classifier layer. In training period, we choose standard Stochastic Gradient Descent (SGD) \cite{krizhevsky2012imagenet} optimizer with a batch size of 2, a momentum of 0.9 and a weight decay of 5e-4. The learning rate is set to 1e-5 initially and follows the poly learning rate policy \cite{chen2017rethinking} with a poly power of 0.9. The total number of training iterations is set to 200000. We set the probability of triggering random mirror and Gaussian blur to 50\%. In addition, the radius in Gaussian blur is randomly selected from the range of {[}0, 1{]}.

\section{Experimental Results\label{sec:Experimental-Results}}

\begin{table}[t]
\caption{The comparison results with DRPC. This table shows the generalization results for SRSS settings from GTA5. The best results of each setting are marked in bold.\label{tab:compare DRPC}}

\begin{centering}
\centering{}\doublerulesep=0.5pt \resizebox{0.44\textwidth}{!}{
\begin{tabular}{ccccc}
\hline 
Network & SRSS Setting & Methods & mIoU & mIoU $\uparrow$\tabularnewline
\hline 
\multirow{12}{*}{ResNet-101} & \multirow{4}{*}{GTA5 to Cityscapes} & Source Only \cite{yue2019domain} & 33.6 & \multirow{2}{*}{8.9}\tabularnewline
 &  & DRPC \cite{yue2019domain} & 42.5 & \tabularnewline
\cline{3-5} 
 &  & Source Only & 34.0 & \multirow{2}{*}{\textbf{9.7}}\tabularnewline
 &  & Ours & \textbf{43.7} & \tabularnewline
\cline{2-5}
 & \multirow{4}{*}{GTA5 to BDDS} & Source Only \cite{yue2019domain} & 27.8 & \multirow{2}{*}{10.9}\tabularnewline
 &  & DRPC \cite{yue2019domain} & 38.7 & \tabularnewline
\cline{3-5}
 &  & Source Only & 28.1 & \multirow{2}{*}{\textbf{11.5}}\tabularnewline
 &  & Ours & \textbf{39.6} & \tabularnewline
\cline{2-5}
 & \multirow{4}{*}{GTA5 to Mapillary} & Source Only \cite{yue2019domain} & 28.3 & \multirow{2}{*}{9.8}\tabularnewline
 &  & DRPC \cite{yue2019domain} & 38.1 & \tabularnewline
\cline{3-5}
 &  & Source Only & 28.6 & \multirow{2}{*}{\textbf{10.5}}\tabularnewline
 &  & Ours & \textbf{39.1} & \tabularnewline
\hline 
\end{tabular}}
\par\end{centering}
\vspace{-4mm}
\end{table}

\begin{table}[t]
\caption{The comparison results with DRPC. This table shows the generalization results for SRSS settings from SYNTHIA. The best results of each setting are marked in bold.\label{tab:compare DRPC-1}}

\begin{centering}
\centering{}\doublerulesep=0.5pt \resizebox{0.44\textwidth}{!}{
\begin{tabular}{ccccc}
\hline 
Network & SRSS Setting & Methods & mIoU & mIoU $\uparrow$\tabularnewline
\hline 
\multirow{12}{*}{ResNet-101} & \multirow{4}{*}{SYNTHIA to Cityscapes} & Source Only \cite{yue2019domain} & 29.7 & \multirow{2}{*}{7.9}\tabularnewline
 &  & DRPC \cite{yue2019domain} & 37.6 & \tabularnewline
\cline{3-5} 
 &  & Source Only & 30.2 & \multirow{2}{*}{\textbf{9.5}}\tabularnewline
 &  & Ours & \textbf{39.7} & \tabularnewline
\cline{2-5}
 & \multirow{4}{*}{SYNTHIA to BDDS} & Source Only \cite{yue2019domain} & 25.6 & \multirow{2}{*}{8.7}\tabularnewline
 &  & DRPC \cite{yue2019domain} & 34.3 & \tabularnewline
\cline{3-5}
 &  & Source Only & 25.9 & \multirow{2}{*}{\textbf{9.5}}\tabularnewline
 &  & Ours & \textbf{35.3} & \tabularnewline
\cline{2-5}
 & \multirow{4}{*}{SYNTHIA to Mapillary} & Source Only \cite{yue2019domain} & 28.7 & \multirow{2}{*}{5.4}\tabularnewline
 &  & DRPC \cite{yue2019domain} & 34.1 & \tabularnewline
\cline{3-5}
 &  & Source Only & 29.5 & \multirow{2}{*}{\textbf{6.9}}\tabularnewline
 &  & Ours & \textbf{36.4} & \tabularnewline
\hline 
\end{tabular}}
\par\end{centering}
\vspace{-4mm}
\end{table}

\subsection{Comparison with State-of-Arts\label{subsec:Comparison-with-State-of-Art}}

In this subsection, extensive experiments are conducted to show the generalization performance of our framework. Since few work has been devoted to Domain Generalization, we compare our results with the only known three Domain Generalization methods: DPRC \cite{yue2019domain}, IBN \cite{pan2018two} and SW \cite{pan2019switchable}. Note that DPRC, IBN and SW are based on different backbone networks, i.e., ResNet-101 (DPRC), ResNet-50 (IBN) and VGG16 (SW). Therefore, we implement our method on three different backbone networks to make a fair comparison.

\textbf{Comparison with DRPC.} Following DRPC \cite{yue2019domain}, we report the results on three different target domains (Cityscapes, BDDS and Mapillary) from two source domains (GTA5 and SYNTHIA). Specifically, we implement our framework in six experimental settings: ``GTA5 to Cityscapes'', ``GTA5 to BDDS'', ``GTA5 to Mapillary'', ``SYNTHIA to Cityscapes'', ``SYNTHIA to BDDS'' and ``SYNTHIA to Mapillary''. DRPC \cite{yue2019domain} introduce many real-wolrd domains in network training with a pyramid consistency regularization. As shown in Tab. \ref{tab:compare DRPC} and Tab. \ref{tab:compare DRPC-1}, we can see that our proposed framework achieves superior results in all settings. In addition, our framework also has a better mIoU gain in each experimental setting. This illustrates that our proposed framework is more capable of improving generalization performance.

\begin{figure*}[tp]
\begin{centering}
\centering{}\centering \resizebox{0.9\textwidth}{!}{ \includegraphics[width=30cm]{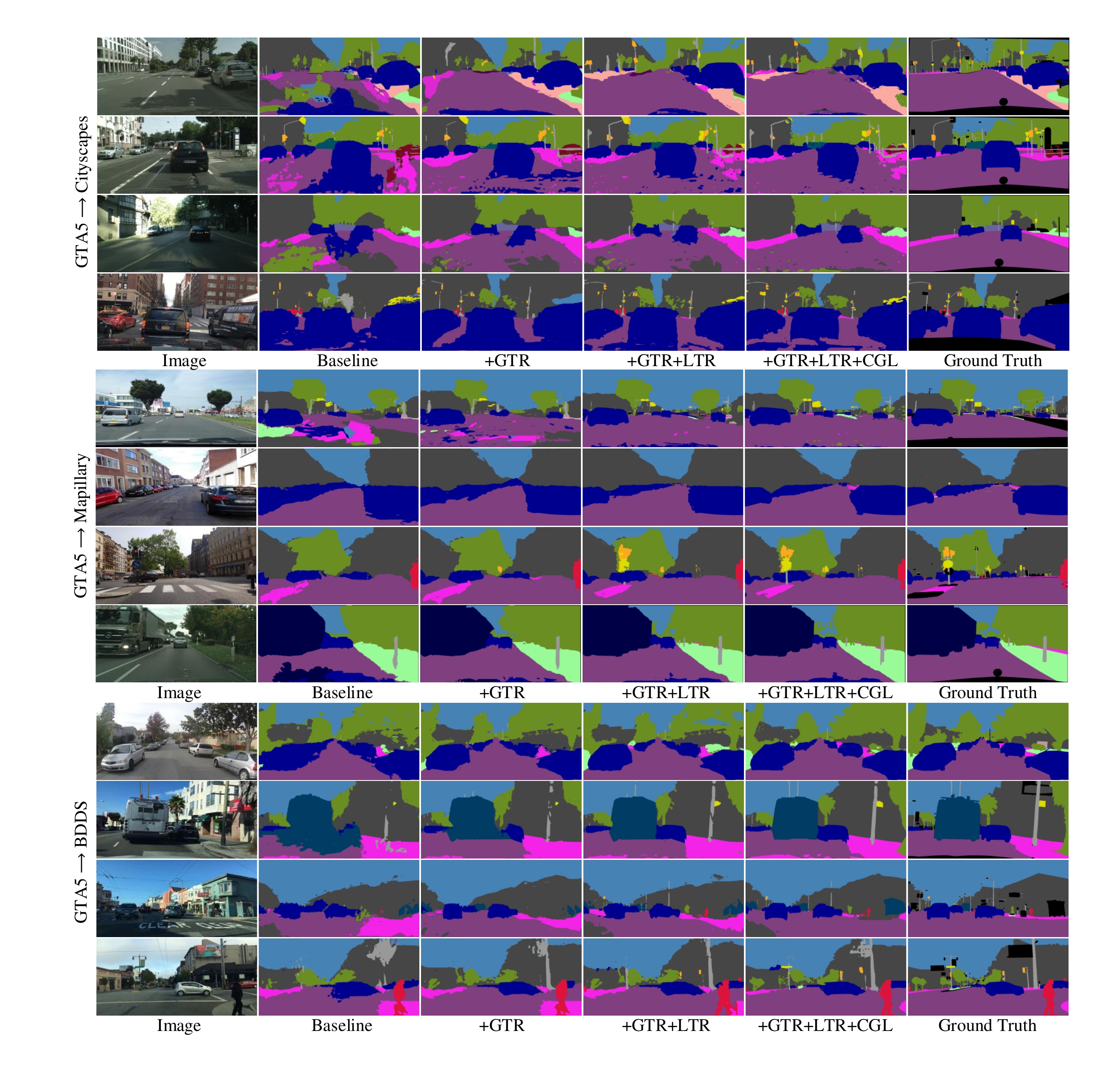}}
\par\end{centering}
\vspace{-4mm}
\caption{Qualitative Domain Generalization based SRSS results from GTA5 to Cityscapes, Mapillary and BDDS. It can be seen that the segmentation performance is improved by adding the proposed modules progressively. Note that all results are obtained under the condition of TCPS. (The figure is best viewed in color)\label{fig:Qualitative-Domain-Generalizatio}}
\vspace{-5mm}
\end{figure*}

\begin{table}[th]
\vspace{-4mm}
\caption{Quantitative results of the comparison with IBN. The best results are marked in bold.\label{tab:compare IBN}}
\begin{centering}
\centering{}\doublerulesep=0.5pt \resizebox{0.44\textwidth}{!}{%
\begin{tabular}{ccccc}
\hline 
Network & SRSS Setting & Methods & mIoU & mIoU $\uparrow$\tabularnewline
\hline 
\multirow{4}{*}{ResNet-50} & \multirow{4}{*}{GTA5 to Cityscapes} & Source Only \cite{pan2018two} & 22.2 & \multirow{2}{*}{\textbf{7.4}}\tabularnewline
 &  & IBN \cite{pan2018two} & 29.6 & \tabularnewline
\cline{3-5}
 &  & Source Only & 31.7 & \multirow{2}{*}{6.9}\tabularnewline
 &  & Ours & \textbf{38.6} & \tabularnewline
\hline 
\end{tabular}}
\par\end{centering}
\vspace{-2mm}
\end{table}

\begin{table}[th]
\caption{Comparison results of ours and combination of ours and IBN. The best results are marked in bold.\label{tab:compare IBN-1}}
\begin{centering}
\centering{}\doublerulesep=0.5pt \resizebox{0.39\textwidth}{!}{%
\begin{tabular}{cccc}
\hline 
Network & SRSS Setting & Methods & mIoU\tabularnewline
\hline 
\multirow{3}{*}{ResNet-50} & \multirow{3}{*}{GTA5 to Cityscapes} & Source Only & 31.7\tabularnewline
 &  & Ours & 38.6\tabularnewline
 &  & Ours+IBN \cite{pan2018two} & \textbf{39.5}\tabularnewline
\hline 
\end{tabular}}
\par\end{centering}
\vspace{-6mm}
\end{table}

\textbf{Comparison with IBN.} We follow the experimental setting (``GTA5 to Cityscapes'') in IBN \cite{pan2018two} to evaluate our framework using backbone network ResNet-50. More specifically, the training set only contains the labeled data in GTA5. Since the ground-truth of testing set in Cityscapes is missing, we use the validation set as the testing set in our experiments. IBN utilizes Instance Normalization (IN) and Batch Normalization (BN) to capture more invariant information from appearance changes while preserving content related information. Our method also learns invariant information, but we devote to data randomization while IBN devotes to network structure design. From the comparison shown in Tab. \ref{tab:compare IBN}, it can be seen that although the mIoU gain of ours is slightly weaker than IBN, our framework achieves a mIoU of 38.6\% which is significantly 9\% better than IBN. In addition, our method is complementary with IBN. To demonstrate that, the additional experiment on model with both IBN and our method is conducted. The results are shown in Tab. \ref{tab:compare IBN-1}. We can see that the model with both IBN and our method boosts the performance to 39.5\%, which is 0.9\% better than the model with only our method.

\begin{table}[th]
\vspace{-4mm}
\caption{Quantitative results of the comparison with SW. The best results are obtained by in-column comparison and marked in bold.\label{tab:compare SW}}
\centering{}\centering{}\doublerulesep=0.5pt \resizebox{0.42\textwidth}{!}{%
\begin{tabular}{ccccc}
\hline 
Network & SRSS Setting & Methods & mIoU & mIoU $\uparrow$\tabularnewline
\hline 
\multirow{4}{*}{VGG-16} & \multirow{4}{*}{GTA5 to Cityscapes} & Source Only & - & \multirow{2}{*}{-}\tabularnewline
 &  & SW \cite{pan2019switchable} & 35.7 & \tabularnewline
\cline{3-5}
 &  & Source Only & 31.4 & \multirow{2}{*}{5.8}\tabularnewline
 &  & Ours & \textbf{37.2} & \tabularnewline
\hline 
\end{tabular}}
\vspace{-2mm}
\end{table}

\textbf{Comparison with SW.} SW \cite{pan2019switchable} is a modified model from IBN \cite{pan2018two}. It provides an integration of several whitening techniques and standardization techniques. SW can adaptively set appropriate whitening or standardization statistics for a specific task. SW offers more possibilities for learning invariant information than IBN. Tab. \ref{tab:compare SW} shows the performance of SW and our method under the experimental setting ``GTA5 to Cityscapes''. With the same backbone network (VGG16), our framework achieves a mIoU of 37.2\% and outperforms the SW model by 1.5\%.

\textbf{Comparison with Domain Adaption methods.} All methods above are conducted with Domain Generalization manner, where the network training has no access to target domain. Now we compare our results with previous state-of-the-art Domain Adaption methods which are target domain-accessible. In view of the most extensive Domain Adaption methods are based on VGG-16 and ResNet-101, we conduct experiments on the same backbone networks. Since most of previous works conducted domain adaption from GTA5/SYNTHIA to Cityscapes, we present the comparison results under the same SRSS settings in Tab. \ref{tab:Experiment-results-of} and Tab. \ref{tab:Experiment-results-of-1}, respectively. It can be seen that in each setting, our method is on par or even better than existing Domain Adaption methods.

\begin{table}[t]
\caption{Experiment results from GTA5 to Cityscapes. The best results are marked in bold. DA and DG denote Domain Adaption and Domain Generalization respectively. \label{tab:Experiment-results-of}}

\centering{}\doublerulesep=0.5pt \resizebox{0.43\textwidth}{!}{
\begin{tabular}{ccccc}
\hline 
Network & DA / DG & Method & Access Tgt & mIoU\tabularnewline
\hline 
\multirow{13}{*}{VGG-16} & \multirow{12}{*}{DA} & FCN wild \cite{hoffman2016fcns} & $\checked$ & 27.1\tabularnewline
 &  & CDA \cite{zhang2017curriculum} & $\checked$ & 28.9\tabularnewline
 &  & CyCADA \cite{hoffman2017cycada} & $\checked$ & 34.8\tabularnewline
 &  & ROAD \cite{chen2018road} & $\checked$ & 35.9\tabularnewline
 &  & I2I \cite{murez2018image} & $\checked$ & 31.8\tabularnewline
 &  & AdaptSegNet \cite{tsai2018learning} & $\checked$ & 35.0\tabularnewline
 &  & SSF-DAN \cite{du2019ssf} & $\checked$ & \textbf{37.7}\tabularnewline
 &  & DCAN \cite{wu2018dcan} & $\checked$ & 36.2\tabularnewline
 &  & CBST \cite{zou2018domain} & $\checked$ & 30.9\tabularnewline
 &  & CLAN \cite{luo2019taking} & $\checked$ & 36.6\tabularnewline
 &  & ADVENT \cite{vu2019advent} & $\checked$ & 36.1\tabularnewline
 &  & DPR \cite{tsai2019domain} & $\checked$ & 37.5\tabularnewline
\cline{2-5}
 & DG & Ours & $\times$ & 37.2\tabularnewline
\hline 
\multirow{9}{*}{ResNet-101} & \multirow{8}{*}{DA} & CyCADA \cite{hoffman2017cycada} & $\checked$ & 42.7\tabularnewline
 &  & ROAD \cite{chen2018road} & $\checked$ & 39.4\tabularnewline
 &  & I2I \cite{murez2018image} & $\checked$ & 35.4\tabularnewline
 &  & AdaptSegNet \cite{tsai2018learning} & $\checked$ & 41.4\tabularnewline
 &  & DCAN \cite{wu2018dcan} & $\checked$ & 41.7\tabularnewline
 &  & CLAN \cite{luo2019taking} & $\checked$ & 43.2\tabularnewline
 &  & ADVENT \cite{vu2019advent} & $\checked$ & 43.8\tabularnewline
 &  & DPR \cite{tsai2019domain} & $\checked$ & \textbf{46.5}\tabularnewline
\cline{2-5}
 & DG & Ours & $\times$ & 43.7\tabularnewline
\hline 
\end{tabular}}

\vspace{-4mm}
\end{table}

\begin{table}[t]
\caption{Experiment results from SYNTHIA to Cityscapes. The best results are marked in bold. DA and DG denote Domain Adaption and Domain Generalization.\label{tab:Experiment-results-of-1}}

\centering{}\doublerulesep=0.5pt \resizebox{0.47\textwidth}{!}{%
\begin{tabular}{ccccc}
\hline 
Network & DA / DG & Method & Access Tgt & mIoU\tabularnewline
\hline 
\multirow{8}{*}{VGG-16} & \multirow{7}{*}{DA} & FCN wild \cite{hoffman2016fcns} & $\checked$ & 20.2\tabularnewline
 &  & CDA \cite{zhang2017curriculum} & $\checked$ & 29.0\tabularnewline
 &  & ROAD \cite{chen2018road} & $\checked$ & \textbf{36.2}\tabularnewline
 &  & DCAN \cite{wu2018dcan} & $\checked$ & 35.4\tabularnewline
 &  & CBST \cite{zou2018domain} & $\checked$ & 35.4\tabularnewline
 &  & ADVENT \cite{vu2019advent} & $\checked$ & 31.4\tabularnewline
 &  & DPR \cite{tsai2019domain} & $\checked$ & 33.7\tabularnewline
\cline{2-5}
 & DG & Ours & $\times$ & 35.8\tabularnewline
\hline 
\multirow{3}{*}{ResNet-101} & \multirow{2}{*}{DA} & ADVENT \cite{vu2019advent} & $\checked$ & \textbf{40.8}\tabularnewline
 &  & DPR \cite{tsai2019domain} & $\checked$ & 40.0\tabularnewline
\cline{2-5}
 & DG & Ours & $\times$ & 39.7\tabularnewline
\hline 
\end{tabular}}

\vspace{-4mm}
\end{table}

\subsection{Ablation Studies\label{sec:Ablation-Study}}

In this subsection, we demonstrate the benefit of each design in our approach. Without loss of generality, we only exhibit the generalization results from GTA5 (G) to Cityscapes (C), BDDS (B) and Mapillary (M) with the backbone network ResNet-101. In Tab. \ref{tab:Performance-contribution-of}, we detail the mIoU improvement by progressively adding components: Global Texture Randomization (GTR), Local Texture Randomization (LTR) and Consistency between GTR and LTR (CGL). Besides, to demonstrate benefits of the proposed Texture Complexity based Painting Selection (TCPS), the same ablation experiments are conducted under different conditions (with or without TCPS).

\begin{table}[t]
\caption{Performance contribution of each proposed design. This evaluation is conducted from GTA5 (G) to Cityscapes (C), BDDS (B) and Mapillary (M). The best results are in bold.\label{tab:Performance-contribution-of}}

\begin{centering}
\centering{}\doublerulesep=0.5pt \resizebox{0.47\textwidth}{!}{
\begin{tabular}{ccccccccc}
\hline 
\multirow{2}{*}{} & \multirow{2}{*}{Methods} & \multirow{2}{*}{TCPS} & \multirow{2}{*}{GTR} & \multirow{2}{*}{LTR} & \multirow{2}{*}{CGL} & \multicolumn{3}{c}{mIoU}\tabularnewline
\cline{7-9}
 &  &  &  &  &  & G$\rightarrow$C & G$\rightarrow$B & G$\rightarrow$M\tabularnewline
\hline 
(a) & ResNet-101 &  &  &  &  & 34.0 & 28.1 & 28.6\tabularnewline
\hline 
(b) & $+$GTR &  & $\checked$ &  &  & 39.2 & 35.4 & 32.1\tabularnewline
(c) & $+$LTR &  &  & $\checked$ &  & 41.8 & 36.1 & 33.7\tabularnewline
(d) & $+$GTR+LTR &  & $\checked$ & $\checked$ &  & 42.4 & 38.6 & 37.8\tabularnewline
(e) & $+$GTR+LTR+CGL &  & $\checked$ & $\checked$ & $\checked$ & 43.2 & 39.1 & 38.4\tabularnewline
\hline 
(f) & $+$TCPS$+$GTR & $\checked$ & $\checked$ &  &  & 40.0 & 36.2 & 33.4\tabularnewline
(g) & $+$TCPS$+$LTR & $\checked$ &  & $\checked$ &  & 42.4 & 36.8 & 34.8\tabularnewline
(h) & $+$TCPS$+$GTR+LTR & $\checked$ & $\checked$ & $\checked$ &  & 43.0 & 39.1 & 38.5\tabularnewline
(i) & $+$TCPS$+$GTR+LTR+CGL & $\checked$ & $\checked$ & $\checked$ & $\checked$ & \textbf{43.7} & \textbf{39.6} & \textbf{39.1}\tabularnewline
\hline 
\end{tabular}}
\par\end{centering}
\vspace{-4mm}
\end{table}

\begin{table}[t]
\caption{The comparisons of LTR with different style combinations. The best results of each SRSS setting are marked in bold.\label{tab:ablation-styles-in-LTR}}

\begin{centering}
\centering{}\doublerulesep=0.5pt \resizebox{0.48\textwidth}{!}{
\begin{tabular}{cccccc}
\hline 
\multirow{2}{*}{} & \multirow{2}{*}{Style Combination} & \multicolumn{4}{c}{mIoU}\tabularnewline
\cline{3-6}
 &  & G$\rightarrow$C & G$\rightarrow$B & G$\rightarrow$M & Avg\tabularnewline
\hline 
\multirow{1}{*}{(a)} & raw style + one painting style & \textbf{43.7} & \textbf{39.6} & \textbf{39.1} & \textbf{40.8}\tabularnewline
\multirow{1}{*}{(b)} & raw style + two painting styles & 42.6 & 37.3 & 37.8 & 39.2\tabularnewline
\multirow{1}{*}{(c)} & raw style + three painting styles & 41.9 & 37.0 & 37.3 & 38.7\tabularnewline
\hline 
\end{tabular}}
\par\end{centering}
\vspace{-4mm}
\end{table}

\begin{table}[t]
\vspace{-4mm}
\caption{The comparisons of LTR with different component proportions. The best results of each SRSS setting are marked in bold.\label{tab:ablation-styles-in-LTR-1}}

\begin{centering}
\centering{}\doublerulesep=0.5pt \resizebox{0.48\textwidth}{!}{%
\begin{tabular}{cccccc}
\hline 
 & Proportion & \multicolumn{4}{c}{mIoU}\tabularnewline
\cline{3-6}
 & (raw style : single painting style) & G$\rightarrow$C & G$\rightarrow$B & G$\rightarrow$M & Avg\tabularnewline
\hline 
(a) & 3 : 1 & 41.7 & 37.3 & 36.8 & 38.6\tabularnewline
(b) & 2 : 1 & 42.4 & 37.8 & 38.1 & 39.4\tabularnewline
(c) & 1 : 1 & \textbf{43.7} & \textbf{39.6} & \textbf{39.1} & \textbf{40.8}\tabularnewline
(d) & 1 : 2 & 42.7 & 37.3 & 37.6 & 39.2\tabularnewline
(e) & 1 : 3 & 42.0 & 36.8 & 37.3 & 38.7\tabularnewline
\hline 
\end{tabular}}
\par\end{centering}
\vspace{-4mm}
\end{table}

\begin{table}[th]
\caption{The comparison between CGL and its derivatives. The best results of each SRSS setting are marked in bold.\label{tab:ablation-consistensy}}

\begin{centering}
\centering{}\doublerulesep=0.5pt \resizebox{0.39\textwidth}{!}{
\begin{tabular}{ccccc}
\hline 
\multirow{2}{*}{Consistency Strategy} & \multicolumn{4}{c}{mIoU}\tabularnewline
\cline{2-5}
 & G$\rightarrow$C & G$\rightarrow$B & G$\rightarrow$M & Avg\tabularnewline
\hline 
CGL & 43.7 & \textbf{39.6} & \textbf{39.1} & \textbf{40.8}\tabularnewline
\multirow{1}{*}{CGS} & 43.4 & 39.2 & 38.8 & 40.5\tabularnewline
\multirow{1}{*}{CLS} & 43.6 & 39.6 & 38.4 & 40.5\tabularnewline
\multirow{1}{*}{CTF} & \textbf{43.8} & 39.2 & 38.9 & 40.6\tabularnewline
\hline 
\end{tabular}}
\par\end{centering}
\vspace{-6mm}
\end{table}

\begin{table}[th]
\vspace{-4mm}
\caption{Performance comparison between unreal paintings and real (or synthetic) images. The best results of each SRSS setting are marked in bold.\label{tab:ablation-painting-image}}

\begin{centering}
\centering{}\doublerulesep=0.5pt \resizebox{0.39\textwidth}{!}{
\begin{tabular}{ccc}
\hline 
SRSS Setting & Type of Style Image & mIoU\tabularnewline
\hline 
\multirow{4}{*}{GTA5 to Cityscapes} & Paintings & \textbf{43.7}\tabularnewline
 & Images from Mapillary & 39.3\tabularnewline
 & Images from BDDS & 40.7\tabularnewline
 & Images from SYNTHIA & 36.8\tabularnewline
\hline 
\multirow{4}{*}{GTA5 to BDDS} & Paintings & \textbf{39.6}\tabularnewline
 & Images from Cityscapes & 37.5\tabularnewline
 & Images from Mapillary & 36.7\tabularnewline
 & Images from SYNTHIA & 33.6\tabularnewline
\hline 
\multirow{4}{*}{GTA5 to Mapillary} & Paintings & \textbf{39.1}\tabularnewline
 & Images from BDDS & 35.5\tabularnewline
 & Images from Cityscapes & 35.8\tabularnewline
 & Images from SYNTHIA & 32.9\tabularnewline
\hline 
\end{tabular}}
\par\end{centering}
\vspace{-4mm}
\end{table}

\textbf{Effects of Global Texture Randomization (GTR).} One can observe that compared to baseline (Tab. \ref{tab:Performance-contribution-of} a), the proposed GTR (Tab. \ref{tab:Performance-contribution-of} b) helps boost the performance from 34.0\% to 39.2\%, from 28.1\% to 35.4\% and from 28.6\% to 32.1\% on GTA5 (G)$\rightarrow$Cityscapes (C), GTA5 (G)$\rightarrow$BDDS (B) and GTA5 (G)$\rightarrow$Mapillary (M). We can also see that the model with GTR can achieve 5.2\% improvement in experimental setting G$\rightarrow$C, 7.3\% improvement on G$\rightarrow$B and 3.5\% improvement on G$\rightarrow$M. The improvements of the GTR clearly demonstrate the benefit of utilizing diverse texture styles.

\begin{figure*}[tp]
\begin{centering}
\centering{}\centering \resizebox{0.9\textwidth}{!}{ \includegraphics[width=30cm]{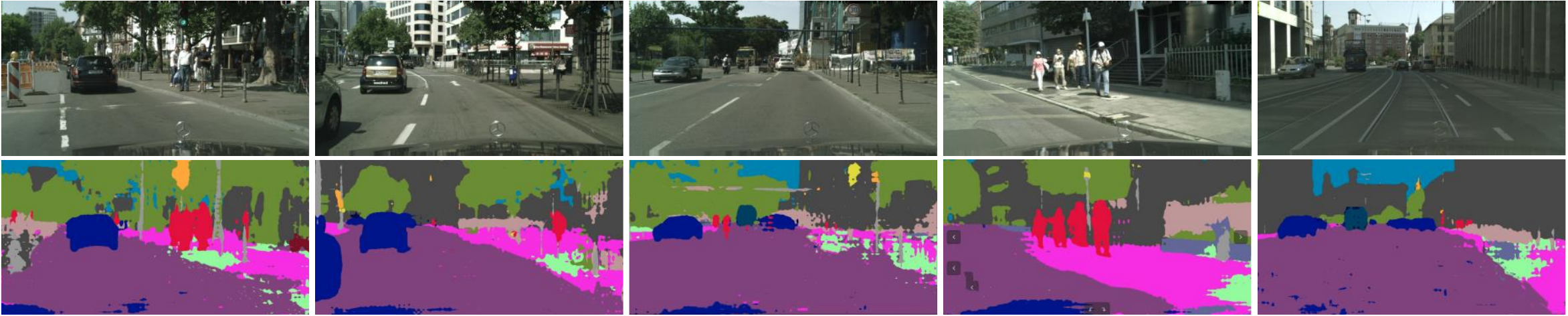}}
\par\end{centering}
\vspace{-2mm}

\caption{Some failure examples of our framework. There is a misclassification when the sidewalk shows a similar appearance with terrain. (The figure is best viewed in color.)\label{fig:Qualitative-Domain-Generalizatio-1}}

\vspace{-4mm}
\end{figure*}

\textbf{Effects of Local Texture Randomization (LTR).} To evaluate the benefits of the proposed module LTR, we also re-implement our approach only with LTR (Tab. \ref{tab:Performance-contribution-of} c). In terms of mIoU, our framework with LTR achieves 41.8\%/36.1\%/33.7\% on G$\rightarrow$C/G$\rightarrow$B/G$\rightarrow$M. Meanwhile,
compared to baseline (Tab. \ref{tab:Performance-contribution-of} a), model with LTR can greatly boost the results by 7.8\%/8\%/5.1\% from GTA5 to Cityscapes/BDDS/Mapillary. According to the comparison between model only with GTR (Tab. \ref{tab:Performance-contribution-of} b) and model only with LTR (Tab. \ref{tab:Performance-contribution-of} c), we can see that the proposed method LTR performs better than GTR in all three settings (G$\rightarrow$C, G$\rightarrow$B, G$\rightarrow$M). We consider to combine GTR and LTR in our approach aiming to lead both global and local texture styles to our network. Compared to (Tab. \ref{tab:Performance-contribution-of} b) and (Tab. \ref{tab:Performance-contribution-of} c) respectively, model adopted GTR and LTR (Tab. \ref{tab:Performance-contribution-of} d) achieves remarkable results and gives a further improvement in all experimental setting. It means our mainly proposed two designs (GTR and LTR) can complement with each other. Moreover, it is important to find out the key factors affecting the performance of LTR. Several experiments are conducted to explore more combinations of panting styles in LTR. As shown in Tab. \ref{tab:ablation-styles-in-LTR}, adding painting styles is unable to boost the performance but obviously degrade it. We observe that adding painting styles also decreases the proportion of the raw style in image. Tab. \ref{tab:ablation-styles-in-LTR-1} shows the ablation results of LTR with different component proportions. It is worthy noting that we utilize the single style image in LTR. Compared to LTR with multiple painting styles (Tab. \ref{tab:ablation-styles-in-LTR} b/c), LTR with single painting style (Tab. \ref{tab:ablation-styles-in-LTR-1} d/e) achieves almost the same performances when the proportion of raw style is same. It illustrates that the proportion of the raw style is the main factor to affect the performance of LTR. As shown in Tab. \ref{tab:ablation-styles-in-LTR}, our approach achieves the best performance when the raw style has the same proportion with painting styles.

\textbf{Effects of Consistency between GTR and LTR (CGL).} When further adding CGL to our framework, the performance of our framework (Tab. \ref{tab:Performance-contribution-of} e) is booted to 43.2\%/39.1\%/38.4\% on G$\rightarrow$C/G$\rightarrow$B/G$\rightarrow$M. Compared to the model with GTR and LTR (Tab. \ref{tab:Performance-contribution-of} d), although the CGL has little improvements, it has a widely stable performance gains in all settings. Besides, we show results for different consistent strategies in Tab. \ref{tab:ablation-consistensy}. The experiments include: Consistency between GTR and Source flow (CGS), Consistency between LTR and Source flow (CLS) and Consistency among Three Flows (CTF). As shown in Tab. \ref{tab:ablation-consistensy}, CGL shows the best performance on most of evaluation metrics.

\textbf{Effects of Texture Complexity based Painting Selection (TCPS). }Ablations with and without TCPS are performed to verify the proposed painting selection method. Note that when not adopting TCPS, 15 paintings are randomly selected from the painting dataset. Ablation results are given by taking the average value of 5 times experiments. The random painting selection of each experiment is independent. Compared with directly random selection from the painting dataset (Tab. \ref{tab:Performance-contribution-of} b-e), the model with TCPS (Tab. \ref{tab:Performance-contribution-of} h-i) improves the performance in each ablation setting by 0.5\%\textasciitilde 1.3\%. This confirms the effectiveness of our TCPS.

\textbf{Unreal Painting vs. Real or Synthetic Image. }In this paper, there are three types of datasets: Unreal Painting (``Painter by Numbers''), Synthetic Image (GTA5 and SYNTHIA) and Real Image (Cityscapes, BDDS and Mapillary). We mainly demonstrate the domain generalization ability of network can be significantly improved with unreal paintings. Whether it can meet the equal level of performance when using real or synthetic images to execute GTR and LTR? As shown in Tab. \ref{tab:ablation-painting-image}, we conduct experiments to address the above concern. It can be seen that the utilization of paintings outperforms using real and synthetic images in all SRSS settings. This is due to unreal paintings of dataset ``Painting by Numbers'' introduce various irregular and indistinct textures, enforcing the network to learn from other domain-invariant cues. While in real or synthetic dataset, the source images are transferred to a new domain. And the performance mainly depends on the similarity between the transferred domain and target domain.

\begin{table}[t]
\caption{Parameter analysis towards the number of training iterations. This evaluation is conducted from GTA5 (G) to Cityscapes (C), BDDS (B) and Mapillary (M). The best results of each SRSS setting are marked in bold.\label{tab:ablation-iteration}}

\begin{centering}
\centering{}\doublerulesep=0.5pt \resizebox{0.39\textwidth}{!}{
\begin{tabular}{cccccccc}
\hline 
\multirow{2}{*}{Iterations} & \multicolumn{7}{c}{mIoU}\tabularnewline
\cline{2-8}
 & G$\rightarrow$C &  & G$\rightarrow$B &  & G$\rightarrow$M &  & Avg\tabularnewline
\hline 
100000 & 38.9 &  & 36.3 &  & 31.3 &  & 35.5\tabularnewline
150000 & 42.0 &  & 37.8 &  & 38.3 &  & 39.4\tabularnewline
200000 & \textbf{43.7} &  & \textbf{39.6} &  & \textbf{39.1} &  & \textbf{40.8}\tabularnewline
250000 & 43.4 &  & 39.1 &  & 38.8 &  & 40.4\tabularnewline
300000 & 43.3 &  & 38.8 &  & 38.7 &  & 40.3\tabularnewline
\hline 
\end{tabular}}
\par\end{centering}
\vspace{-6mm}
\end{table}

\begin{table}[t]
\caption{Parameter analysis of $\lambda$ in Equation \ref{eq:}. The remarkable results are marked in blue while the rest are marked in green.\label{tab:ablation-lamda}}

\begin{centering}
\centering{}\doublerulesep=0.5pt \resizebox{0.41\textwidth}{!}{
\begin{tabular}{ccccccccc}
\hline 
\multirow{2}{*}{$\lambda$} &  & \multicolumn{7}{c}{mIoU}\tabularnewline
\cline{2-9}
 &  & G$\rightarrow$C &  & G$\rightarrow$B &  & G$\rightarrow$M &  & Avg\tabularnewline
\hline 
2 &  & 40.3 &  & 36.3 &  & 33.4 &  & \textcolor{green}{36.7}\tabularnewline
4 &  & 41.9 &  & 38.4 &  & 37.2 &  & \textcolor{blue}{39.2}\tabularnewline
6 &  & 42.7 &  & 38.6 &  & 37.5 &  & \textcolor{blue}{39.6}\tabularnewline
8 &  & 43.2 &  & 39.0 &  & 38.1 &  & \textcolor{blue}{40.1}\tabularnewline
10 &  & 43.4 &  & 39.3 &  & 38.7 &  & \textcolor{blue}{40.5}\tabularnewline
12 &  & 43.1 &  & 38.9 &  & 38.3 &  & \textcolor{blue}{40.1}\tabularnewline
14 &  & 43.0 &  & 38.9 &  & 38.1 &  & \textcolor{blue}{40.0}\tabularnewline
16 &  & 42.8 &  & 38.5 &  & 37.7 &  & \textcolor{blue}{39.7}\tabularnewline
18 &  & 40.6 &  & 36.5 &  & 33.7 &  & \textcolor{green}{36.9}\tabularnewline
20 &  & 40.1 &  & 36.4 &  & 33.5 &  & \textcolor{green}{36.7}\tabularnewline
\hline 
\end{tabular}}
\par\end{centering}
\vspace{-4mm}
\end{table}

\begin{table}[t]
\caption{Parameter analysis of $\beta$ in Equation \ref{eq:beta}.The best results of each SRSS setting are marked in bold.\label{tab:ablation-beta}}

\begin{centering}
\centering{}\doublerulesep=0.5pt \resizebox{0.42\textwidth}{!}{
\begin{tabular}{ccccccccc}
\hline 
\multirow{2}{*}{$\beta$} &  & \multicolumn{7}{c}{mIoU}\tabularnewline
\cline{2-9}
 &  & G$\rightarrow$C &  & G$\rightarrow$B &  & G$\rightarrow$M &  & Avg\tabularnewline
\hline 
1e-3 &  & 42.8 &  & 39.3 &  & 36.9 &  & 39.6\tabularnewline
1e-4 &  & 43.1 &  & 39.4 &  & 38.7 &  & 40.4\tabularnewline
1e-5 &  & \textbf{43.7} &  & \textbf{39.6} &  & \textbf{39.1} &  & \textbf{40.8}\tabularnewline
1e-6 &  & 42.9 &  & 39.5 &  & 38.8 &  & 40.4\tabularnewline
1e-7 &  & 42.5 &  & 39.1 &  & 37.4 &  & 39.7\tabularnewline
\hline 
\end{tabular}}
\par\end{centering}
\vspace{-5mm}
\end{table}

\textbf{Analysis of Parameters.} The performances of our approach with different training iterations are shown in Tab. \ref{tab:ablation-iteration}. It can be seen that setting the number to 200000 produces the best results. Applying a larger or smaller iteration number leads to a degradation of the domain generalization performance. The $\lambda$ in Equation \ref{eq:} is the parameter to control the standard deviation of Gaussian kernel. We evaluate the performance of our method to this parameter. The results are recorded in Tab. \ref{tab:ablation-lamda}. For scores which are marked in blue, we can see that any two neighboring results are extremely close to each other (within 1\%). In addition, all blue scores show outstanding performance and surpass the green scores in a large margin. The scores marked in blue are results of $\lambda$ varying from 4 to 16. In order to increase the diversity of paintings while ensuring the performance, we randomly choose $\lambda$ from $[4,16]$ in each iteration. The $\beta$ in Equation \ref{eq:beta} is the parameter to trade off the quality of segmentation loss with consistency loss. As shown in Tab. \ref{tab:ablation-beta}, we can see that when $\beta=1e-5$, our method achieves the best performance. The threshold of texture complexity $\mathcal{T}_{exc}$ has the same analysis trends with $\lambda$. From Tab. \ref{tab:ablation-texture-complexity}, one can observe that when $\mathcal{T}_{exc}\in[0.55,0.65]$, our method achieves outstanding performances.

\begin{table}[t]
\caption{Parameter analysis towards texture complexity $\mathcal{T}_{exc}$ of selected paintings. \label{tab:ablation-texture-complexity}}

\begin{centering}
\centering{}\doublerulesep=0.5pt \resizebox{0.41\textwidth}{!}{
\begin{tabular}{ccccccccc}
\hline 
\multirow{2}{*}{$\mathcal{T}_{exc}$} &  & \multicolumn{7}{c}{mIoU}\tabularnewline
\cline{2-9}
 &  & G$\rightarrow$C &  & G$\rightarrow$B &  & G$\rightarrow$M &  & Avg\tabularnewline
\hline 
0.45 &  & 41.6 &  & 37.3 &  & 36.7 &  & \textcolor{green}{38.5}\tabularnewline
0.5 &  & 42.4 &  & 38.0 &  & 36.9 &  & \textcolor{green}{39.1}\tabularnewline
0.55 &  & 43.0 &  & 39.0 &  & 38.2 &  & \textcolor{blue}{40.1}\tabularnewline
0.6 &  & 43.4 &  & 39.3 &  & 38.8 &  & \textcolor{blue}{40.5}\tabularnewline
0.65 &  & 43.2 &  & 39.2 &  & 38.5 &  & \textcolor{blue}{40.3}\tabularnewline
0.7 &  & 41.8 &  & 38.2 &  & 36.6 &  & \textcolor{green}{38.9}\tabularnewline
0.75 &  & 41.0 &  & 36.8 &  & 35.7 &  & \textcolor{green}{37.8}\tabularnewline
\hline 
\end{tabular}}
\par\end{centering}
\vspace{-5mm}
\end{table}

\textbf{Quantitative Results.} Fig. \ref{fig:Qualitative-Domain-Generalizatio} provides some visual examples of the semantic segmentation results under three Cross-Domain settings. We can see that the proposed framework (the 5-th column) can generates segmentation with more details (e.g., pole, traffic sign and traffic light). Comparing results in the 2-nd column and the 3-rd column, the model with GTR produces less noisy regions, especially in the category of road. In addition, the usage of LTR brings a better boundary decision in various settings. For example, segmentation boundaries of car, pole, bicycle, traffic light and traffic sigh in the 4-th column are clearer than the first few columns. Fig. \ref{fig:Qualitative-Domain-Generalizatio-1} shows some failure examples. One can observe that when the sidewalk has a dark green texture appearance, our method may misclassify some regions of sidewalk into terrain. The example in the 2-nd column shows that the shadow in sidewalk are misclassified, but human can easily recognize the shadow. This is due to that our model can not completely eliminate the effect of texture even using GTR and LTR.

\section{Conclusion\label{sec:Conclusion}}

In this paper, we mainly present two texture randomization mechanisms for Domain Generalization. To relieve the network's strong bias towards recognizing texture, we propose a mechanism named Global Texture Randomization (GTR) to learn domain-invariant information. Next, Local Texture Randomization (LTR) mechanism is proposed to address the cases where only local regions are texture-different between the synthetic and real-world images.Moreover, to harmonize the two randomization mechanisms, consistency between GTR and LTR (CGL) is introduced. Besides, Texture Complexity based Painting Selection (TCPS) is proposed to ensure paintings are reliable enough for the above texture randomization mechanisms. Extensive experiments indicate that our approach achieves the state-of-the-art results in multiple synthetic-to-real dataset settings with different network backbones, which clearly demonstrates the effectiveness of our method.

{\small{}\bibliographystyle{IEEEtran}
\bibliography{reference}

\begin{thebibliography}{10}
\providecommand{\url}[1]{#1}
\csname url@samestyle\endcsname
\providecommand{\newblock}{\relax}
\providecommand{\bibinfo}[2]{#2}
\providecommand{\BIBentrySTDinterwordspacing}{\spaceskip=0pt\relax}
\providecommand{\BIBentryALTinterwordstretchfactor}{4}
\providecommand{\BIBentryALTinterwordspacing}{\spaceskip=\fontdimen2\font plus
\BIBentryALTinterwordstretchfactor\fontdimen3\font minus
  \fontdimen4\font\relax}
\providecommand{\BIBforeignlanguage}[2]{{%
\expandafter\ifx\csname l@#1\endcsname\relax
\typeout{** WARNING: IEEEtran.bst: No hyphenation pattern has been}%
\typeout{** loaded for the language `#1'. Using the pattern for}%
\typeout{** the default language instead.}%
\else
\language=\csname l@#1\endcsname
\fi
#2}}
\providecommand{\BIBdecl}{\relax}
\BIBdecl

\bibitem{zhang2019cascaded}
P.~Zhang, W.~Liu, Y.~Lei, H.~Lu, and X.~Yang, ``Cascaded context pyramid for
  full-resolution 3d semantic scene completion,'' in \emph{Proceedings of the
  International Conference on Computer Vision (ICCV)}.\hskip 1em plus 0.5em
  minus 0.4em\relax IEEE, 2019, pp. 7801--7810.

\bibitem{long2015fully}
J.~Long, E.~Shelhamer, and T.~Darrell, ``Fully convolutional networks for
  semantic segmentation,'' in \emph{Proceedings of the Computer Vision and
  Pattern Recognition (CVPR)}.\hskip 1em plus 0.5em minus 0.4em\relax IEEE,
  2015, pp. 3431--3440.

\bibitem{chen2017deeplab}
L.~Chen, G.~Papandreou, I.~Kokkinos, K.~Murphy, and A.~L. Yuille, ``Deeplab:
  Semantic image segmentation with deep convolutional nets, atrous convolution,
  and fully connected crfs,'' \emph{Transactions on Pattern Analysis and
  Machine Intelligence (TPAMI)}, vol.~40, no.~4, pp. 834--848, 2017.

\bibitem{he2017mask}
K.~He, G.~Gkioxari, P.~Doll{\'a}r, and R.~Girshick, ``Mask r-cnn,'' in
  \emph{Proceedings of the International Conference on Computer Vision
  (ICCV)}.\hskip 1em plus 0.5em minus 0.4em\relax IEEE, 2017, pp. 2961--2969.

\bibitem{handa2015scenenet}
A.~Handa, V.~Patraucean, V.~Badrinarayanan, S.~Stent, and R.~Cipolla,
  ``Scenenet: understanding real world indoor scenes with synthetic data. arxiv
  preprint (2015),'' \emph{arXiv preprint arXiv:1511.07041}, 2015.

\bibitem{richter2016playing}
S.~R. Richter, V.~Vineet, S.~Roth, and V.~Koltun, ``Playing for data: Ground
  truth from computer games,'' in \emph{Proceedings of the European Conference
  on Computer Vision (ECCV)}.\hskip 1em plus 0.5em minus 0.4em\relax Springer,
  2016, pp. 102--118.

\bibitem{hoffman2017cycada}
J.~Hoffman, E.~Tzeng, T.~Park, J.~Zhu, P.~Isola, K.~Saenko, A.~A. Efros, and
  T.~Darrell, ``Cycada: Cycle-consistent adversarial domain adaptation,''
  \emph{arXiv preprint arXiv:1711.03213}, 2017.

\bibitem{hoffman2016fcns}
J.~Hoffman, D.~Wang, F.~Yu, and T.~Darrell, ``Fcns in the wild: Pixel-level
  adversarial and constraint-based adaptation,'' \emph{arXiv preprint
  arXiv:1612.02649}, 2016.

\bibitem{zhang2017curriculum}
Y.~Zhang, P.~David, and B.~Gong, ``Curriculum domain adaptation for semantic
  segmentation of urban scenes,'' in \emph{Proceedings of the International
  Conference on Computer Vision (ICCV)}.\hskip 1em plus 0.5em minus 0.4em\relax
  IEEE, 2017, pp. 2020--2030.

\bibitem{li2018learning}
D.~Li, Y.~Yang, Y.~Song, and T.~M. Hospedales, ``Learning to generalize:
  Meta-learning for domain generalization,'' in \emph{Thirty-Second AAAI
  Conference on Artificial Intelligence}, 2018.

\bibitem{balaji2018metareg}
Y.~Balaji, S.~Sankaranarayanan, and R.~Chellappa, ``Metareg: Towards domain
  generalization using meta-regularization,'' in \emph{Advances in Neural
  Information Processing Systems (NuerIPS)}, 2018, pp. 998--1008.

\bibitem{ghifary2015domain}
M.~Ghifary, W.~Bastiaan~Kleijn, M.~Zhang, and D.~Balduzzi, ``Domain
  generalization for object recognition with multi-task autoencoders,'' in
  \emph{Proceedings of the International Conference on Computer Vision (ICCV)},
  2015, pp. 2551--2559.

\bibitem{tobin2017domain}
J.~Tobin, R.~Fong, A.~Ray, J.~Schneider, W.~Zaremba, and P.~Abbeel, ``Domain
  randomization for transferring deep neural networks from simulation to the
  real world,'' in \emph{2017 IEEE/RSJ international conference on intelligent
  robots and systems (IROS)}.\hskip 1em plus 0.5em minus 0.4em\relax IEEE,
  2017, pp. 23--30.

\bibitem{GeirhosImageNet}
R.~Geirhos, P.~Rubisch, C.~Michaelis, M.~Bethge, F.~A. Wichmann, and
  W.~Brendel, ``Imagenet-trained cnns are biased towards texture; increasing
  shape bias improves accuracy and robustness,'' \emph{arXiv preprint
  arXiv:1811.12231}, 2018.

\bibitem{bousmalis2017unsupervised}
K.~Bousmalis, N.~Silberman, D.~Dohan, D.~Erhan, and D.~Krishnan, ``Unsupervised
  pixel-level domain adaptation with generative adversarial networks,'' in
  \emph{Proceedings of the Computer Vision and Pattern Recognition
  (CVPR)}.\hskip 1em plus 0.5em minus 0.4em\relax IEEE, 2017, pp. 3722--3731.

\bibitem{kolmogorov2004energy}
V.~Kolmogorov and R.~Zabin, ``What energy functions can be minimized via graph
  cuts?'' \emph{Transactions on Pattern Analysis and Machine Intelligence
  (TPAMI)}, vol.~26, no.~2, pp. 147--159, 2004.

\bibitem{meltzer2005globally}
T.~Meltzer, C.~Yanover, and Y.~Weiss, ``Globally optimal solutions for energy
  minimization in stereo vision using reweighted belief propagation,'' in
  \emph{Proceedings of the International Conference on Computer Vision (ICCV)},
  vol.~1.\hskip 1em plus 0.5em minus 0.4em\relax IEEE, 2005, pp. 428--435.

\bibitem{kolmogorov2014new}
V.~Kolmogorov, ``A new look at reweighted message passing,'' \emph{Transactions
  on Pattern Analysis and Machine Intelligence (TPAMI)}, vol.~37, no.~5, pp.
  919--930, 2014.

\bibitem{grady2004multi}
L.~Grady and G.~Funka-Lea, ``Multi-label image segmentation for medical
  applications based on graph-theoretic electrical potentials,'' in
  \emph{Proceedings of the European Conference on Computer Vision Workshops
  (ECCV Workshops)}.\hskip 1em plus 0.5em minus 0.4em\relax Springer, 2004, pp.
  230--245.

\bibitem{sinop2007seeded}
A.~K. Sinop and L.~Grady, ``A seeded image segmentation framework unifying
  graph cuts and random walker which yields a new algorithm,'' in
  \emph{Proceedings of the International Conference on Computer Vision
  (ICCV)}.\hskip 1em plus 0.5em minus 0.4em\relax IEEE, 2007, pp. 1--8.

\bibitem{dong2015sub}
X.~Dong, J.~Shen, L.~Shao, and L.~Van~Gool, ``Sub-markov random walk for image
  segmentation,'' \emph{Transactions on Image Processing (TIP)}, vol.~25,
  no.~2, pp. 516--527, 2015.

\bibitem{blake2011markov}
A.~Blake, P.~Kohli, and C.~Rother, \emph{Markov random fields for vision and
  image processing}.\hskip 1em plus 0.5em minus 0.4em\relax MIT press, 2011.

\bibitem{shen2019submodular}
J.~Shen, X.~Dong, J.~Peng, X.~Jin, L.~Shao, and F.~Porikli, ``Submodular
  function optimization for motion clustering and image segmentation,''
  \emph{Transactions on Neural Networks and Learning Systems (TNNLS)}, vol.~30,
  no.~9, pp. 2637--2649, 2019.

\bibitem{shen2017higher}
J.~Shen, J.~Peng, X.~Dong, L.~Shao, and F.~Porikli, ``Higher order energies for
  image segmentation,'' \emph{Transactions on Image Processing (TIP)}, vol.~26,
  no.~10, pp. 4911--4922, 2017.

\bibitem{yu2015multi}
F.~Yu and V.~Koltun, ``Multi-scale context aggregation by dilated
  convolutions,'' \emph{arXiv preprint arXiv:1511.07122}, 2015.

\bibitem{chen2014semantic}
L.-C. Chen, G.~Papandreou, I.~Kokkinos, K.~Murphy, and A.~L. Yuille, ``Semantic
  image segmentation with deep convolutional nets and fully connected crfs,''
  \emph{arXiv preprint arXiv:1412.7062}, 2014.

\bibitem{liu2015semantic}
Z.~Liu, X.~Li, P.~Luo, C.-C. Loy, and X.~Tang, ``Semantic image segmentation
  via deep parsing network,'' in \emph{Proceedings of the International
  Conference on Computer Vision (ICCV)}.\hskip 1em plus 0.5em minus 0.4em\relax
  IEEE, 2015, pp. 1377--1385.

\bibitem{zheng2015conditional}
S.~Zheng, S.~Jayasumana, B.~Romera-Paredes, V.~Vineet, Z.~Su, D.~Du, C.~Huang,
  and P.~H. Torr, ``Conditional random fields as recurrent neural networks,''
  in \emph{Proceedings of the International Conference on Computer Vision
  (ICCV)}.\hskip 1em plus 0.5em minus 0.4em\relax IEEE, 2015, pp. 1529--1537.

\bibitem{peng2017large}
C.~Peng, X.~Zhang, G.~Yu, G.~Luo, and J.~Sun, ``Large kernel matters--improve
  semantic segmentation by global convolutional network,'' in \emph{Proceedings
  of the Computer Vision and Pattern Recognition (CVPR)}.\hskip 1em plus 0.5em
  minus 0.4em\relax IEEE, 2017, pp. 4353--4361.

\bibitem{pang2019towards}
Y.~Pang, Y.~Li, J.~Shen, and L.~Shao, ``Towards bridging semantic gap to
  improve semantic segmentation,'' in \emph{Proceedings of the International
  Conference on Computer Vision (ICCV)}.\hskip 1em plus 0.5em minus 0.4em\relax
  IEEE, 2019, pp. 4230--4239.

\bibitem{huang2019see}
Y.~Huang, Q.~Wang, W.~Jia, and X.~He, ``See more than once--kernel-sharing
  atrous convolution for semantic segmentation,'' \emph{arXiv preprint
  arXiv:1908.09443}, 2019.

\bibitem{chen2018road}
Y.~Chen, W.~Li, and L.~Van~Gool, ``Road: Reality oriented adaptation for
  semantic segmentation of urban scenes,'' in \emph{Proceedings of the Computer
  Vision and Pattern Recognition (CVPR)}.\hskip 1em plus 0.5em minus
  0.4em\relax IEEE, 2018, pp. 7892--7901.

\bibitem{long2016unsupervised}
M.~Long, H.~Zhu, J.~Wang, and M.~I. Jordan, ``Unsupervised domain adaptation
  with residual transfer networks,'' in \emph{Advances in Neural Information
  Processing Systems (NuerIPS)}, 2016, pp. 136--144.

\bibitem{ganin2014unsupervised}
Y.~Ganin and V.~Lempitsky, ``Unsupervised domain adaptation by
  backpropagation,'' \emph{arXiv preprint arXiv:1409.7495}, 2014.

\bibitem{cariucci2017autodial}
F.~M. Cariucci, L.~Porzi, B.~Caputo, E.~Ricci, and S.~R. Bul{\`o}, ``Autodial:
  Automatic domain alignment layers,'' in \emph{Proceedings of the
  International Conference on Computer Vision (ICCV)}.\hskip 1em plus 0.5em
  minus 0.4em\relax IEEE, 2017, pp. 5077--5085.

\bibitem{chen2017no}
Y.~Chen, W.~Chen, Y.~Chen, B.~Tsai, Y.~Frank~Wang, and M.~Sun, ``No more
  discrimination: Cross city adaptation of road scene segmenters,'' in
  \emph{Proceedings of the International Conference on Computer Vision
  (ICCV)}.\hskip 1em plus 0.5em minus 0.4em\relax IEEE, 2017, pp. 1992--2001.

\bibitem{tsai2018learning}
Y.~Tsai, W.~Hung, S.~Schulter, K.~Sohn, M.~Yang, and M.~Chandraker, ``Learning
  to adapt structured output space for semantic segmentation,'' in
  \emph{Proceedings of the Computer Vision and Pattern Recognition
  (CVPR)}.\hskip 1em plus 0.5em minus 0.4em\relax IEEE, 2018, pp. 7472--7481.

\bibitem{tzeng2017adversarial}
E.~Tzeng, J.~Hoffman, K.~Saenko, and T.~Darrell, ``Adversarial discriminative
  domain adaptation,'' in \emph{Proceedings of the Computer Vision and Pattern
  Recognition (CVPR)}.\hskip 1em plus 0.5em minus 0.4em\relax IEEE, 2017, pp.
  7167--7176.

\bibitem{luo2019taking}
Y.~Luo, L.~Zheng, T.~Guan, J.~Yu, and Y.~Yang, ``Taking a closer look at domain
  shift: Category-level adversaries for semantics consistent domain
  adaptation,'' in \emph{Proceedings of the Computer Vision and Pattern
  Recognition (CVPR)}.\hskip 1em plus 0.5em minus 0.4em\relax IEEE, 2019, pp.
  2507--2516.

\bibitem{vu2019advent}
T.~Vu, H.~Jain, M.~Bucher, M.~Cord, and P.~P{\'e}rez, ``Advent: Adversarial
  entropy minimization for domain adaptation in semantic segmentation,'' in
  \emph{Proceedings of the Computer Vision and Pattern Recognition
  (CVPR)}.\hskip 1em plus 0.5em minus 0.4em\relax IEEE, 2019, pp. 2517--2526.

\bibitem{tsai2019domain}
Y.~Tsai, K.~Sohn, S.~Schulter, and M.~Chandraker, ``Domain adaptation for
  structured output via discriminative patch representations,'' in
  \emph{Proceedings of the International Conference on Computer Vision
  (ICCV)}.\hskip 1em plus 0.5em minus 0.4em\relax IEEE, 2019, pp. 1456--1465.

\bibitem{zhu2017unpaired}
J.~Zhu, T.~Park, P.~Isola, and A.~A. Efros, ``Unpaired image-to-image
  translation using cycle-consistent adversarial networks,'' in
  \emph{Proceedings of the International Conference on Computer Vision
  (ICCV)}.\hskip 1em plus 0.5em minus 0.4em\relax IEEE, 2017, pp. 2223--2232.

\bibitem{murez2018image}
Z.~Murez, S.~Kolouri, D.~Kriegman, R.~Ramamoorthi, and K.~Kim, ``Image to image
  translation for domain adaptation,'' in \emph{Proceedings of the Computer
  Vision and Pattern Recognition (CVPR)}.\hskip 1em plus 0.5em minus
  0.4em\relax IEEE, 2018, pp. 4500--4509.

\bibitem{zou2018domain}
Y.~Zou, Z.~Yu, B.~Kumar, and J.~Wang, ``Domain adaptation for semantic
  segmentation via class-balanced self-training,'' \emph{arXiv preprint
  arXiv:1810.07911}, 2018.

\bibitem{wu2018dcan}
Z.~Wu, X.~Han, Y.-L. Lin, M.~Gokhan~Uzunbas, T.~Goldstein, S.~Nam~Lim, and
  L.~S. Davis, ``Dcan: Dual channel-wise alignment networks for unsupervised
  scene adaptation,'' in \emph{Proceedings of the European Conference on
  Computer Vision (ECCV)}, 2018, pp. 518--534.

\bibitem{du2019ssf}
L.~Du, J.~Tan, H.~Yang, J.~Feng, X.~Xue, Q.~Zheng, X.~Ye, and X.~Zhang,
  ``Ssf-dan: Separated semantic feature based domain adaptation network for
  semantic segmentation,'' in \emph{Proceedings of the International Conference
  on Computer Vision (ICCV)}.\hskip 1em plus 0.5em minus 0.4em\relax IEEE,
  2019, pp. 982--991.

\bibitem{muandet2013domain}
K.~Muandet, D.~Balduzzi, and B.~Sch{\"o}lkopf, ``Domain generalization via
  invariant feature representation,'' in \emph{International Conference on
  Machine Learning (ICML)}, 2013, pp. 10--18.

\bibitem{gan2016learning}
C.~Gan, T.~Yang, and B.~Gong, ``Learning attributes equals multi-source domain
  generalization,'' in \emph{Proceedings of the Computer Vision and Pattern
  Recognition (CVPR)}.\hskip 1em plus 0.5em minus 0.4em\relax IEEE, 2016, pp.
  87--97.

\bibitem{gong2013reshaping}
B.~Gong, K.~Grauman, and F.~Sha, ``Reshaping visual datasets for domain
  adaptation,'' in \emph{Advances in Neural Information Processing Systems
  (NuerIPS)}, 2013, pp. 1286--1294.

\bibitem{li2018domain}
H.~Li, S.~Jialin~Pan, S.~Wang, and A.~C. Kot, ``Domain generalization with
  adversarial feature learning,'' in \emph{Proceedings of the Computer Vision
  and Pattern Recognition (CVPR)}.\hskip 1em plus 0.5em minus 0.4em\relax IEEE,
  2018, pp. 5400--5409.

\bibitem{li2017deeper}
D.~Li, Y.~Yang, Y.-Z. Song, and T.~M. Hospedales, ``Deeper, broader and artier
  domain generalization,'' in \emph{Proceedings of the International Conference
  on Computer Vision (ICCV)}.\hskip 1em plus 0.5em minus 0.4em\relax IEEE,
  2017, pp. 5542--5550.

\bibitem{yue2019domain}
X.~Yue, Y.~Zhang, S.~Zhao, A.~Sangiovanni-Vincentelli, K.~Keutzer, and B.~Gong,
  ``Domain randomization and pyramid consistency: Simulation-to-real
  generalization without accessing target domain data,'' in \emph{Proceedings
  of the International Conference on Computer Vision (ICCV)}, 2019, pp.
  2100--2110.

\bibitem{pan2018two}
X.~Pan, P.~Luo, J.~Shi, and X.~Tang, ``Two at once: Enhancing learning and
  generalization capacities via ibn-net,'' in \emph{Proceedings of the European
  Conference on Computer Vision (ECCV)}, 2018, pp. 464--479.

\bibitem{pan2019switchable}
X.~Pan, X.~Zhan, J.~Shi, X.~Tang, and P.~Luo, ``Switchable whitening for deep
  representation learning,'' in \emph{Proceedings of the International
  Conference on Computer Vision (ICCV)}.\hskip 1em plus 0.5em minus 0.4em\relax
  IEEE, 2019, pp. 1863--1871.

\bibitem{huang2018decorrelated}
L.~Huang, D.~Yang, B.~Lang, and J.~Deng, ``Decorrelated batch normalization,''
  in \emph{Proceedings of the Computer Vision and Pattern Recognition (CVPR)},
  2018, pp. 791--800.

\bibitem{li2017universal}
Y.~Li, C.~Fang, J.~Yang, Z.~Wang, X.~Lu, and M.-H. Yang, ``Universal style
  transfer via feature transforms,'' in \emph{Advances in Neural Information
  Processing Systems (NuerIPS)}, 2017, pp. 386--396.

\bibitem{ulyanov2016instance}
D.~Ulyanov, A.~Vedaldi, and V.~Lempitsky, ``Instance normalization: The missing
  ingredient for fast stylization,'' \emph{arXiv preprint arXiv:1607.08022},
  2016.

\bibitem{ioffe2015batch}
S.~Ioffe and C.~Szegedy, ``Batch normalization: Accelerating deep network
  training by reducing internal covariate shift,'' \emph{arXiv preprint
  arXiv:1502.03167}, 2015.

\bibitem{ciregan2012multi}
D.~Ciregan, U.~Meier, and J.~Schmidhuber, ``Multi-column deep neural networks
  for image classification,'' in \emph{Proceedings of the Computer Vision and
  Pattern Recognition (CVPR)}.\hskip 1em plus 0.5em minus 0.4em\relax IEEE,
  2012, pp. 3642--3649.

\bibitem{sato2015apac}
I.~Sato, H.~Nishimura, and K.~Yokoi, ``Apac: Augmented pattern classification
  with neural networks,'' \emph{arXiv preprint arXiv:1505.03229}, 2015.

\bibitem{wan2013regularization}
L.~Wan, M.~Zeiler, S.~Zhang, Y.~Le~Cun, and R.~Fergus, ``Regularization of
  neural networks using dropconnect,'' in \emph{International Conference on
  Machine Learning (ICML)}, 2013, pp. 1058--1066.

\bibitem{simard2003best}
P.~Y. Simard, D.~Steinkraus, J.~C. Platt \emph{et~al.}, ``Best practices for
  convolutional neural networks applied to visual document analysis.'' in
  \emph{Icdar}, vol.~3, no. 2003, 2003.

\bibitem{lemley2017smart}
J.~Lemley, S.~Bazrafkan, and P.~Corcoran, ``Smart augmentation learning an
  optimal data augmentation strategy,'' \emph{Ieee Access}, vol.~5, pp.
  5858--5869, 2017.

\bibitem{tran2017bayesian}
T.~Tran, T.~Pham, G.~Carneiro, L.~Palmer, and I.~Reid, ``A bayesian data
  augmentation approach for learning deep models,'' in \emph{Advances in Neural
  Information Processing Systems (NuerIPS)}, 2017, pp. 2797--2806.

\bibitem{devries2017dataset}
T.~DeVries and G.~W. Taylor, ``Dataset augmentation in feature space,''
  \emph{arXiv preprint arXiv:1702.05538}, 2017.

\bibitem{dreossi2018counterexample}
T.~Dreossi, S.~Ghosh, X.~Yue, K.~Keutzer, A.~Sangiovanni-Vincentelli, and S.~A.
  Seshia, ``Counterexample-guided data augmentation,'' \emph{arXiv preprint
  arXiv:1805.06962}, 2018.

\bibitem{zhang2017mixup}
H.~Zhang, M.~Cisse, Y.~N. Dauphin, and D.~Lopez-Paz, ``mixup: Beyond empirical
  risk minimization,'' \emph{arXiv preprint arXiv:1710.09412}, 2017.

\bibitem{devries2017cutout}
T.~DeVries and G.~W. Taylor, ``Improved regularization of convolutional neural
  networks with cutout,'' \emph{arXiv preprint arXiv:1708.04552}, 2017.

\bibitem{yun2019cutmix}
S.~Yun, D.~Han, S.~J. Oh, S.~Chun, J.~Choe, and Y.~Yoo, ``Cutmix:
  Regularization strategy to train strong classifiers with localizable
  features,'' in \emph{Proceedings of the International Conference on Computer
  Vision (ICCV)}.\hskip 1em plus 0.5em minus 0.4em\relax IEEE, 2019, pp.
  6023--6032.

\bibitem{kothe2003edge}
U.~K{\"o}the, ``Edge and junction detection with an improved structure
  tensor,'' in \emph{Joint Pattern Recognition Symposium (JPRS)}.\hskip 1em
  plus 0.5em minus 0.4em\relax Springer, 2003, pp. 25--32.

\bibitem{huang2017arbitrary}
X.~Huang and S.~Belongie, ``Arbitrary style transfer in real-time with adaptive
  instance normalization,'' in \emph{Proceedings of the International
  Conference on Computer Vision (ICCV)}.\hskip 1em plus 0.5em minus 0.4em\relax
  IEEE, 2017, pp. 1501--1510.

\bibitem{french2019semi}
G.~French, T.~Aila, S.~Laine, M.~Mackiewicz, and G.~Finlayson,
  ``Semi-supervised semantic segmentation needs strong, high-dimensional
  perturbations,'' \emph{arXiv preprint arXiv:1906.01916}, 2019.

\bibitem{andrews1998special}
L.~C. Andrews, \emph{Special functions of mathematics for engineers}.\hskip 1em
  plus 0.5em minus 0.4em\relax Spie Press, 1998, vol.~49.

\bibitem{simonyan2014very}
K.~Simonyan and A.~Zisserman, ``Very deep convolutional networks for
  large-scale image recognition,'' \emph{arXiv preprint arXiv:1409.1556}, 2014.

\bibitem{he2016deep}
K.~He, X.~Zhang, S.~Ren, and J.~Sun, ``Deep residual learning for image
  recognition,'' in \emph{Proceedings of the Computer Vision and Pattern
  Recognition (CVPR)}.\hskip 1em plus 0.5em minus 0.4em\relax IEEE, 2016, pp.
  770--778.

\bibitem{paszke2019pytorch}
A.~Paszke \emph{et~al.}, ``Pytorch: An imperative style, high-performance deep
  learning library,'' in \emph{Proc. Adv. Neural Inf. Process. Syst.
  (NuerIPS)}, 2019, pp. 8026--8037.

\bibitem{everingham2015pascal}
M.~Everingham, S.~A. Eslami, L.~Van~Gool, C.~K. Williams, J.~Winn, and
  A.~Zisserman, ``The pascal visual object classes challenge: A
  retrospective,'' \emph{International Journal of Computer Vision (IJCV)}, vol.
  111, no.~1, pp. 98--136, 2015.

\bibitem{zhao2017pyramid}
H.~Zhao, J.~Shi, X.~Qi, X.~Wang, and J.~Jia, ``Pyramid scene parsing network,''
  in \emph{Proceedings of the Computer Vision and Pattern Recognition
  (CVPR)}.\hskip 1em plus 0.5em minus 0.4em\relax IEEE, 2017, pp. 2881--2890.

\bibitem{chen2019domain}
M.~Chen, H.~Xue, and D.~Cai, ``Domain adaptation for semantic segmentation with
  maximum squares loss,'' in \emph{Proceedings of the International Conference
  on Computer Vision (ICCV)}.\hskip 1em plus 0.5em minus 0.4em\relax IEEE,
  2019, pp. 2090--2099.

\bibitem{deng2009imagenet:}
J.~Deng, W.~Dong, R.~Socher, L.-J. Li, K.~Li, and L.~Fei-Fei, ``Imagenet: A
  large-scale hierarchical image database,'' in \emph{Proceedings of the
  Computer Vision and Pattern Recognition (CVPR)}.\hskip 1em plus 0.5em minus
  0.4em\relax IEEE, 2009, pp. 248--255.

\bibitem{krizhevsky2012imagenet}
A.~Krizhevsky, I.~Sutskever, and G.~E. Hinton, ``Imagenet classification with
  deep convolutional neural networks,'' in \emph{Advances in Neural Information
  Processing Systems (NuerIPS)}, 2012, pp. 1097--1105.

\bibitem{chen2017rethinking}
L.~Chen, G.~Papandreou, F.~Schroff, and H.~Adam, ``Rethinking atrous
  convolution for semantic image segmentation,'' \emph{arXiv preprint
  arXiv:1706.05587}, 2017.

\end{thebibliography}
}{\small\par}

\end{document}